\documentclass[sigconf,natbib=true,screen]{acmart}
\usepackage{xspace,balance,tabularx,multirow}
\usepackage{flushend}
\usepackage{tikz}
\usepackage{pgfplots}
\pgfplotsset{compat=1.16}
\usetikzlibrary{patterns}
\usepackage{subfig}
\usepackage[ruled, vlined, linesnumbered]{algorithm2e}
\usepackage{colortbl}
\usepackage{bbold}
\SetKwComment{Comment}{$\triangleright$\ }{}
\usepackage{enumitem}
\usepackage{tablefootnote}
\usepackage{upgreek,textgreek}
\usepackage{pifont}
\usepackage[noabbrev]{cleveref}
\usepackage{titlecaps}
\usepackage{lipsum}
\usepackage{svg}

\pgfplotsset{every tick label/.append style={font=\tiny}}

\newlength{\starsize}
\newlength{\starspread}
\tikzset{starsize/.code={\setlength{\starsize}{#1}},
         starspread/.code={\setlength{\starspread}{#1}}}
\tikzset{starsize=1mm,
         starspread=3mm}
\pgfdeclarepatternformonly[\starspread,\starsize]
  {my fivepointed stars}
  {\pgfpointorigin}
  {\pgfqpoint{\starspread}{\starspread}}
  {\pgfqpoint{\starspread}{\starspread}}
  {
   \pgftransformshift{\pgfqpoint{\starsize}{\starsize}}
   \pgfpathmoveto{\pgfqpointpolar{18}{\starsize}}
   \pgfpathlineto{\pgfqpointpolar{162}{\starsize}}
   \pgfpathlineto{\pgfqpointpolar{306}{\starsize}}
   \pgfpathlineto{\pgfqpointpolar{90}{\starsize}}
   \pgfpathlineto{\pgfqpointpolar{234}{\starsize}}
   \pgfpathclose%
   \pgfusepath{fill}
  }

\makeatletter
\newcommand*\bigcdot{\mathpalette\bigcdot@{.5}}
\newcommand*\bigcdot@[2]{\mathbin{\vcenter{\hbox{\scalebox{#2}{$\m@th#1\bullet$}}}}}
\makeatother

\newcommand{\stitle}[1]{\vspace*{0.5em}\noindent{\bf #1.\/}}

\newcommand{\bigzero}{\mbox{\normalfont\Large\bfseries 0}}
\newcommand{\rvline}{\hspace*{-\arraycolsep}\vline\hspace*{-\arraycolsep}}

\newcommand{\eat}[1]{}

\newcommand{\U}{\mathcal{U}\xspace}
\newcommand{\V}{\mathcal{V}\xspace}
\newcommand{\G}{\mathcal{G}\xspace}
\newcommand{\N}{\mathcal{N}\xspace}
\newcommand{\EDG}{\mathcal{E}\xspace}
\newcommand{\PEDG}{\mathcal{E}^{(+)}\xspace}
\newcommand{\NEDG}{\mathcal{E}^{(-)}\xspace}
\newcommand{\PNGH}{\mathcal{N}^{(+)}\xspace}
\newcommand{\NNGH}{\mathcal{N}^{(-)}\xspace}

\newcommand{\WM}{\mathbf{W}\xspace}
\newcommand{\AM}{\mathbf{A}\xspace}
\newcommand{\PAM}{\mathbf{A}^{(+)}\xspace}
\newcommand{\NAM}{\mathbf{A}^{(-)}\xspace}
\newcommand{\DM}{\mathbf{D}\xspace}
\newcommand{\IM}{\mathbf{I}\xspace}

\newcommand{\MM}{\mathbf{M}\xspace}
\newcommand{\PM}{\mathbf{P}\xspace}

\newcommand{\YM}{\mathbf{Y}\xspace}
\newcommand{\XM}{\mathbf{X}\xspace}
\newcommand{\LM}{\mathbf{L}\xspace}
\newcommand{\UM}{\mathbf{U}\xspace}
\newcommand{\VM}{\mathbf{V}\xspace}
\newcommand{\HM}{\mathbf{H}\xspace}
\newcommand{\ZM}{\mathbf{Z}\xspace}

\newcommand{\BM}{\mathbf{B}\xspace}

\newcommand{\PsiM}{\boldsymbol{\Psi}\xspace}
\newcommand{\PhiM}{\boldsymbol{\Phi}\xspace}

\newcommand{\algo}{\texttt{GegenNet}\xspace}

\newenvironment{customlegend}[1][]{%
    \begingroup
    \csname pgfplots@init@cleared@structures\endcsname
    \pgfplotsset{#1}%
}{%
    \csname pgfplots@createlegend\endcsname
    \endgroup
}%

\def\addlegendimage{\csname pgfplots@addlegendimage\endcsname}

\makeatletter
\newcommand\footnoteref[1]{\protected@xdef\@thefnmark{\ref{#1}}\@footnotemark}
\makeatother

\let\oldnl\nl
\newcommand{\nonl}{\renewcommand{\nl}{\let\nl\oldnl}}

\DeclareMathOperator{\Tr}{\textsf{trace}}

\SetKwComment{Comment}{/* }{ */}

\makeatletter 
\g@addto@macro{\@algocf@init}{\SetKwInOut{Parameter}{Parameters}} 
\makeatother

\definecolor{myred}{HTML}{fd7f6f}
\definecolor{myred_new}{HTML}{D8D8D8}
\definecolor{myred_new2}{HTML}{D7191C}
\definecolor{myblue}{HTML}{7eb0d5}
\definecolor{mygreen}{HTML}{b2e061}
\definecolor{mypurple}{HTML}{bd7ebe}
\definecolor{myorange}{HTML}{ffb55a}
\definecolor{myyellow}{HTML}{ffee65}
\definecolor{mypurple2}{HTML}{beb9db}
\definecolor{mypink}{HTML}{fdcce5}
\definecolor{mycyan}{HTML}{8bd3c7}

\definecolor{myblue2}{HTML}{115f9a}
\definecolor{myred2}{HTML}{c23728}

\definecolor{B0}{HTML}{3C2F80}
\definecolor{B1}{HTML}{012030}
\definecolor{B2}{HTML}{0162A7}
\definecolor{B3}{HTML}{36A7CF}
\definecolor{B4}{HTML}{9AEBA3}
\definecolor{B5}{HTML}{DAFDBA}
\definecolor{B6}{HTML}{45C4B0}

\definecolor{O1}{HTML}{F29E38}
\definecolor{O2}{HTML}{F28444}
\definecolor{O3}{HTML}{D53E0F}

\definecolor{R1}{HTML}{F2889B}

\AtBeginDocument{%
  \providecommand\BibTeX{{%
    \normalfont B\kern-0.5em{\scshape i\kern-0.25em b}\kern-0.8em\TeX}}}

\copyrightyear{2025} 
\acmYear{2025} 
\setcopyright{cc}
\setcctype{by}
\acmConference[CIKM '25]{Proceedings of the 34th ACM International Conference on Information and Knowledge Management}{November 10--14, 2025}{Seoul, Republic of Korea}
\acmBooktitle{Proceedings of the 34th ACM International Conference on Information and Knowledge Management (CIKM '25), November 10--14, 2025, Seoul, Republic of Korea}\acmDOI{10.1145/3746252.3761129}
\acmISBN{979-8-4007-2040-6/2025/11}

\begin{document}

\title{\algo: Spectral Convolutional Neural Networks for Link Sign Prediction in Signed Bipartite Graphs}

\author{Hewen Wang}
\authornote{Both authors contributed equally to the paper.}
\affiliation{%
  \institution{National University of Singapore}
  \country{Singapore}
}
\email{wanghewen@u.nus.edu}
\orcid{0000-0002-9757-4347}

\author{Renchi Yang}
\authornotemark[1]
\affiliation{%
  \institution{Hong Kong Baptist University}
  \country{Hong Kong SAR, China}
}
\email{renchi@hkbu.edu.hk}
\orcid{0000-0002-7284-3096}

\author{Xiaokui Xiao}
\affiliation{%
  \institution{National University of Singapore}
  \country{Singapore}
}
\email{xkxiao@nus.edu.sg}
\orcid{0000-0003-0914-4580}

\renewcommand{\shortauthors}{Trovato and Tobin, et al.}

\begin{abstract}
Given a {\em signed bipartite graph} (SBG) $\G$ with two disjoint node sets $\U$ and $\V$, the goal of link sign prediction is to predict the signs of potential links connecting $\U$ and $\V$ based on known positive and negative edges in $\G$. The majority of existing solutions towards link sign prediction mainly focus on {\em unipartite} signed graphs, which are sub-optimal due to the neglect of node heterogeneity and unique bipartite characteristics of SBGs.
To this end, recent studies adapt {\em graph neural networks} to SBGs by introducing message-passing schemes for both inter-partition ($\U\times\V$) and intra-partition ($\U\times\U$ or $\V\times\V$) node pairs.
However, the fundamental spectral convolutional operators were originally designed for positive links in unsigned graphs, and thus, are not optimal for inferring missing positive or negative links from known ones in SBGs.

Motivated by this, this paper proposes \algo, a novel and effective spectral convolutional neural network model for link sign prediction in SBGs. In particular, \algo achieves enhanced model capacity and high predictive accuracy through three main technical contributions: (i) fast and theoretically grounded spectral decomposition techniques for node feature initialization; (ii) a new spectral graph filter based on the Gegenbauer polynomial basis; and (iii) multi-layer sign-aware spectral convolutional networks alternating Gegenbauer polynomial filters with positive and negative edges.
Our extensive empirical studies reveal that \algo can achieve significantly superior performance (up to a gain of $4.28\%$ in AUC and $11.69\%$ in F1) in link sign prediction compared to 11 strong competitors over 6 benchmark SBG datasets.  
\end{abstract}




\maketitle

\section{Introduction}
{\em Signed bipartite graphs} (SBGs) are an expressive data structure for modeling positive and negative interactions between two heterogeneous sets of real-world entities. Such graph data is prevalent in practical domains such as recommendation systems, sentiment analysis, and opinion mining, where users may express positive or negative sentiments toward items or other entities. On top of that, SBGs enable the modeling of complex social and economic interactions, such as trust/distrust relationships in e-commerce platforms or like/dislike actions in social networks. 
A fundamental task in SBGs is to predict the signs of potential links, i.e., {\em link sign prediction}, which finds numerous applications in online social media~\cite{leskovec2010signed}, recommender systems~\cite{massa2007trust}, and so on.
Distinct from traditional link prediction in general graphs, link sign prediction in SBGs additionally poses challenges in coping with node heterogeneity, lopsided structure, and the presence of positive and negative links.

In the literature~\cite{zhang2024signed}, the majority of existing works~\cite{Wang2017SignedNE,10.1007/978-3-319-93037-4_13,10.1007/978-3-030-16142-2_7,kim2018side,10.1145/3366423.3380038,pmlr-v89-cucuringu19a,Wang2017SignedNE,derr2018signed,Li2020,Huang2019,Huang2021a,doi:10.1137/1.9781611977172.28} towards link sign prediction primarily focus on unipartite signed graphs and resort to learning node representations using approaches based on Skip-gram models~\cite{Mikolov2013} (e.g., DeepWalk, node2vec) or {\em graph neural networks} (GNNs)~\cite{kipf2017semi}.
As reviewed in Section~\ref{sec:SGRL}, most recent advances~\cite{zhang2024signed}  are GNN-based, all of which strongly rely on the message-passing paradigm~\cite{huang2023node,wang2021approximate} or attention mechanisms that are originally dedicated for unsigned unipartite networks that solely consist of positive edges.
Moreover, the design of most of them follows principles including balance theory~\cite{cartwright1956structural} and status theory~\cite{leskovec2010signed}, where the former postulates that ``the friend of my friend is my friend'' and ``the enemy of my enemy is my friend'', while the latter explains social hierarchies and authority relationships.
However, the direct adoption of these techniques for SBGs is inappropriate since an SBG consists of two heterogeneous sets of nodes and positive/negative links connecting them, which render traditional principles insufficient to capture the complex relations and interplay among the nodes.

Recently, a few attempts have been made to address the unique challenges of link sign prediction in SBGs. More precisely, \citet{huang2021signed} extend the balance theory to SBGs through the inter- and intra-partition perspectives and re-design the message functions to aggregate information from different sets (i.e., inter- and intra-partition) of neighborhoods.
To deal with noisy interactions present in real SBGs, \citet{zhang2023contrastive} introduce the contrastive learning mechanism to learn robust node representations, in which a multi-perspective contrastive loss is optimized over augmented graphs created from the inter- and intra-partition perspectives.
By modeling explicit and implicit relations between the inter- and intra-partition nodes, these two approaches are enabled to capture node heterogeneity and bipartite structure in SBGs, and hence, achieve encouraging performance in predicting link signs.
However, akin to their unipartite counterparts, these models are sub-optimal as the adopted spectral convolutional (message-passing) operations are catered for positive edges and depend on the monomial bases. There is still a lack of discussion on the design choices of such foundational operators for both positive and negative links in SBGs.

In response, in this paper, we present \algo, an effective spectral convolutional neural network specifically designed for link sign prediction in SBGs. 
To deepen the understanding of the relations between positive and negative links in SBGs, we recast the link sign prediction task into four curving fitting problems with known and missing links based on our theoretical transformations.
Our empirical observations on real SBGs manifest that existing graph filters based on monomial bases often struggle with capturing both low- and high-frequency spectral signals in SBGs, which motivates the design of our novel spectral graph filter built on the Gegenbauer polynomial basis.
Based thereon, \algo stacks multiple layers of sign-aware spectral convolutional operators with positive and negative links severally using various Gegenbauer polynomial bases, leading to a sophisticated combination that greatly enhances model capacity as pinpointed by our theoretical analysis.
Further, instead of randomly initializing node features as in prior works, which engenders suboptimal feature representations and hinders the effective utilization of graph structure, \algo includes efficacious spectral decomposition techniques that allow us to efficiently construct high-quality initial node embeddings encoding inter- and intra-partition relations between nodes. In summary, our major contributions of this paper are:
\begin{itemize}[leftmargin=*]
\item \textbf{Spectral Feature Initialization for SBGs}:
We develop an effective feature initialization method that can accurately encode structural semantics in SBGs into node embeddings through fast and theoretically grounded spectral decompositions.
\item \textbf{New Spectral Graph Filter for SBGs}: We design a new spectral graph filter based on the Gegenbauer polynomial basis, which overcomes the limitations of previous graph filters.
\item \textbf{Sign-aware Spectral Convolutional Layers for SBGs}: 
We propose \algo that comprises multiple layers of spectral convolutional neural networks using Gegenbauer polynomial filters over positive and negative links in a separate, alternate, and unified manner for stronger model capacity.
\item \textbf{Comprehensive Evaluation}: Extensive experiments comparing \algo against 11 baselines are conducted on 6 real SBG datasets, 
exhibiting the remarkable superiority of \algo over the state of the art in link sign prediction.
\end{itemize}

\section{Related Work}
\subsection{Bipartite Graph Representation Learning}
Existing works for bipartite graph representation learning can be categorized into proximity-preserving and message-passing approaches, where the former seeks to capture both local and global proximity among nodes~\cite{yang2024efficient}, while the latter applies the message-passing paradigm in GNNs to bipartite networks.

More specifically, BiNE \cite{10.1145/3209978.3209987} extends random walk-based Skip-gram models~\cite{Mikolov2013} to non-attributed bipartite graphs.
BiANE \cite{Huang2020a} leverages autoencoders to integrate structural and attribute information by considering intra- and inter-domain proximities severally.
FOBE and HOBE \cite{sybrandt2020fobehobefirsthighorder} preserve type-specific semantic information in bipartite graphs by sampling nodes using different proximities. 
GEBE \cite{yang2022scalable} decomposes proximity matrices capturing multi-hop connectivity among homogeneous and heterogeneous nodes. 
EAGLE \cite{wang2024effective} leverages the factorized feature propagation (FFP) scheme to incorporate long-range dependencies of edges/features.

Amid message-passing-based methods, Cascade-BGNN \cite{Zhu2020b} combines a self-supervised framework with inter- and intra-domain aggregation mechanisms. BiGI \cite{10.1145/3437963.3441783} applies a subgraph-level attention mechanism to maximize the mutual information between local and global node embeddings. DualHGCN \cite{10.1145/3442381.3449954} transforms multiplex bipartite networks into homogeneous hypergraphs and utilizes spectral convolutional operators to capture within- and cross-domain information. Lastly, AnchorGNN \cite{10.14778/3626292.3626300} proposes a global-local learning framework with an anchor-based message-passing schema~\cite{zhou2023slotgat}.

\subsection{Signed Graph Representation Learning}\label{sec:SGRL}
Unlike traditional graph representation learning methods for unsigned graphs, {\em signed graph representation learning} (SGRL) encodes nodes and edges into low-dimensional representations in the presence of the co-existence of positive ties (e.g., trust and friendship) and negative ties (e.g., distrust and enmity). 
Existing works towards SGRL can be mainly categorized into two classes: random walk-based and GNN-based methods.

In particular, random walk-based approaches generate node sequences by traversing the graph and input such sequences into Skip-Gram models~\cite{Mikolov2013} for representation learning. These methods incorporate the signs of edges to capture both positive and negative relationships.
Representative methods include SNE \cite{Wang2017SignedNE}, SIGNet \cite{10.1007/978-3-319-93037-4_13}, and SIDE \cite{kim2018side}.
Signed GNNs extend classic GNN architectures to handle signed graphs by incorporating balance and status theories into the message-passing process. 
SGCN \cite{derr2018signed} utilizes dual representations to handle balanced and unbalanced relationships. SNEA \cite{Li2020} applies attention mechanisms for weighted aggregation of neighborhood information, while SiGAT \cite{Huang2019} incorporates motif-based attention to capture structural properties of signed networks. More models can be found in a recent review~\cite{zhang2024signed}.

Recently, several GNN models~\cite{huang2021signed,zhang2023contrastive} specially designed for SBGs have been proposed. More concretely, SBGNN \cite{huang2021signed} incorporates the balance theory and bipartite nature into the message-passing scheme.
SBGCL \cite{zhang2023contrastive} resorts to contrastive learning over perturbed graphs constructed from the SBGs from the perspectives of inter- and intra-partition.

\section{Preliminaries}

\subsection{Symbol and and Terminology}
Let $\G=(\U,\V,\PEDG,\NEDG)$ be a {\em signed bipartite graph} (SBG), wherein $\U$ and $\V$ denote two disjoint sets of nodes, respectively, $\PEDG$ and $\NEDG$ consist of positive and negative links (a.k.a. edges) between $\U$ and $\V$, respectively, and $\PEDG\cap \NEDG = \emptyset$. We use $\PAM\in \mathbb{R}^{|\U|\times |\V|}$ and $\NAM \in \mathbb{R}^{|\U|\times |\V|}$ to symbolize the positive and negative {\em bi-adjacency} matrix of $\G$, respectively. For each link $e_{u,v}\in \PEDG$ (resp. $e_{u,v}\in \NEDG$), $\PAM_{u,v}=1$ (resp. $\NAM_{u,v}=1$). The complete bi-adjacency matrix of $\G$ is represented by $\AM = \PAM+\NAM$.
Accordingly, the positive, negative, and complete {\em adjacency} matrix of $\G$ can be represented by $\footnotesize\textstyle \tilde{\AM}^{+} = \begin{pmatrix}
  \begin{matrix}
\bigzero
  \end{matrix}
  & \rvline & \PAM \\
\hline
  (\PAM)^{\top} & \rvline &
  \begin{matrix}
\bigzero
  \end{matrix}
\end{pmatrix}$, 
$\footnotesize\textstyle \tilde{\AM}^{-} = \begin{pmatrix}
  \begin{matrix}
\bigzero
  \end{matrix}
  & \rvline & \NAM \\
\hline
  (\NAM)^{\top} & \rvline &
  \begin{matrix}
\bigzero
  \end{matrix}
\end{pmatrix}$, and 
$\footnotesize\textstyle \tilde{\AM} = \begin{pmatrix}
  \begin{matrix}
\bigzero
  \end{matrix}
  & \rvline & \AM \\
\hline
  \AM^{\top} & \rvline &
  \begin{matrix}
\bigzero
  \end{matrix}
\end{pmatrix}\in \mathbb{R}^{(|\U|+|\V|)\times (|\U|+|\V|)}$, respectively. We use $\hat{\AM}^{+}$, $\hat{\AM}^{-}$, and $\hat{\AM}$ to represent their normalized versions, respectively.

\stitle{Problem Statement} Given an SBG $\G=(\U,\V,\PEDG,\NEDG)$, {\em link sign prediction} problem aims at predicting if a pair of nodes $(u_i,v_j)\in \U\times \V$ will be disconnected or connected by a positive or negative link. Mathematically, for any possible link $(u_i,v_j)$, its goal is to find a mapping function $F(u_i,v_j) \rightarrow \{1,-1\}$.

\begin{table}[!t]
\centering
\caption{Classic Link Prediction Functions }
\label{tab:linkpred}
\vspace{-3mm}
\begin{tabular}{lcc}
    \hline
    Name & Matrix Form  & $f(\boldsymbol{\Lambda})$ \\
    \hline
    Common Neighbors & $\AM^2$ & $\boldsymbol{\Lambda}^2$ \\
    $K$-hop RW~\cite{lovasz1993random} & $\hat{\AM}^K$ & $\boldsymbol{\Lambda}^K$ \\
    PPR~\cite{jeh2003scaling}  & $\sum_{k=0}^{\infty}{(1-\alpha)\alpha^k \hat{\AM}^k}$ & \(\sum_{k=0}^\infty {\alpha^k}\boldsymbol{\Lambda}^k\) \\
    HKPR~\cite{chung2007heat}  & $\sum_{k=0}^{\infty}{\frac{e^{-\alpha}\alpha^k}{k!} \hat{\AM}^k}$ & \(\sum_{k=0}^\infty \frac{e^{-\alpha}\alpha^k}{k!}\boldsymbol{\Lambda}^k\) \\
    \hline
\end{tabular}
\vspace{-2ex}
\end{table}

\subsection{Link Prediction Heuristics}\label{sec:linpred}
A common way for link predictions is to calculate the proximity $p(v_i,v_j)$ for any node pair $(v_i,v_j)$ over the input graph~\cite{liben2003link}.
Canonical proximity measures in general (unsigned) graphs can be classified into two major categories: {\em local} and {\em global} heuristics~\cite{lu2011link}, where the former primarily rely on common direct neighbors for measuring node proximity, while the latter exploit the {\em high-order} topological connectivity between nodes. The representative of local heuristics is the number of common neighbors:
\begin{equation*}
p(v_i,v_j) = |\N(v_i) \cap \N(v_j)| = \AM^2_{i,j},
\end{equation*}
where $\N(v_i)$ denotes the number of neighbors of $v_i$ and $\AM$ is the symmetric adjacency matrix of the graph.
Classic global heuristics include the $k$-hop random walks ($k$-hop RW) ~\cite{lovasz1993random}, {\em personalized PageRank} (PPR)~\cite{jeh2003scaling}, and {\em heat kernel PageRank} (HKPR)~\cite{chung2007heat}, etc., whose matrix forms are displayed in Table~\ref{tab:linkpred}, where $\hat{\AM}$ represents the normalized adjacency matrix of the graph\footnote{In standard PPR and HKPR, $\hat{\AM}$ is usually the transition matrix of the graph.} and $\hat{\AM}_{i,j}$ signifies the probability of a random walk from $v_i$ visits $v_j$ at $k$-th step.

\begin{proposition}\label{prop:linpred}
The proximity matrices of common neighbors, $k$-hop RW, PPR, and HKPR can be reformulated into
\begin{equation}\label{eq:linkpred-uniform}
\UM f(\boldsymbol{\Lambda}) \UM^{\top},
\end{equation}
where $f(\cdot)$ is a function defined as in Table~\ref{tab:linkpred}, matrix $\UM$ and diagonal matrix $\boldsymbol{\Lambda}$ contain the eigenvectors and eigenvalues of $\AM$ for common neighbors and of $\hat{\AM}$ for other proximity measures.
\end{proposition}

Our theoretical analyses Proportition~\ref{prop:linpred}\footnote{All missing proofs appear in Appendix~\ref{sec:proof}.} reveal that these heuristics can be unified into a spectral decomposition.

\subsection{Spectral Graph Convolution}\label{sec:spectral-conv}
Let $\hat{\LM}=\IM-\hat{\AM}$ be the {\em normalized} graph Laplacian of a general (unsigned) graph and $\UM\boldsymbol{\Lambda}\UM^\top$ be the eigendecomposition of the normalized adjacency matrix $\hat{\AM}$, where the graph spectrum $\boldsymbol{\Lambda}=\textsf{diag}(\lambda_1,\ldots,\lambda_n)$ is the diagonal matrix of its eigenvalues $-1\le \lambda_1 \le \ldots \le \lambda_n \le 1$ and the columns in $\UM$ are the corresponding eigenvectors.

The graph Fourier transform of an input signal $\mathbf{x}\in \mathbb{R}^n$ is defined as $\UM^\top\mathbf{x}$ and the inverse Fourier transformation is $\UM\mathbf{x}$~\cite{shuman2013emerging}. The spectral graph convolution between a signal $\mathbf{x}$ and a filter $\mathbf{g}$ can be generally expressed by
\begin{equation*}
\mathbf{g} * \mathbf{x} = \UM ( (\UM^{\top}\mathbf{g}) \odot (\UM^\top\mathbf{x}) ) = \UM \mathbf{G} \UM^\top \mathbf{x},\ 
\end{equation*}
where $\mathbf{G}=\textsf{diag}(\UM^\top\mathbf{g})$ is the graph filter in the spectral domain and can be parameterized with any model. 
Since the direct eigendecomposition is immensely time-consuming, it is suggested to approximate $\mathbf{G}$ by a $K$-order polynomial in existing studies~\cite{defferrard2016convolutional,wang2021approximate}: 
\begin{small}
\begin{equation}
\textstyle \mathbf{G} \approx g_{\boldsymbol{\theta}}(\IM-\boldsymbol{\Lambda}) = \sum_{k=0}^{K}{\boldsymbol{\theta}_k P_k(\IM-\boldsymbol{\Lambda})},
\end{equation}
\end{small}
where $g_{\boldsymbol{\theta}}:[0,2]\rightarrow \mathbb{R}$ is a spectral filtering function parameterized by polynomial coefficients $\boldsymbol{\theta}$, $g_{\boldsymbol{\theta}}(\IM-\boldsymbol{\Lambda})=\textsf{diag}(g_{\boldsymbol{\theta}}(1-\lambda_1),\ldots,g_{\boldsymbol{\theta}}(1-\lambda_n))$, and $P_{k}(\cdot)$ stands for an arbitrary polynomial basis. 
Accordingly, the spectral graph convolution can be rewritten as
$\UM g_{\boldsymbol{\theta}}(\IM-\boldsymbol{\Lambda}) \UM^\top \mathbf{x} = 
g_{\boldsymbol{\theta}}(\UM(\IM-\boldsymbol{\Lambda})\UM^\top) \mathbf{x} = 
g_{\boldsymbol{\theta}}(\hat{\LM}) \mathbf{x} = \sum_{k=0}^{K}{\boldsymbol{\theta}_k P_k(\hat{\LM})\mathbf{x}}$, where $\sum_{k=0}^{K}{\boldsymbol{\theta}_k P_k(\hat{\LM})}$ is referred to as the {\em spectral graph filter}.

\subsection{Graph Neural Networks}
The majority of existing GNNs essentially apply spectral graph convolution over input node attributes $\XM$. For instance, the node representations $\ZM$ obtained via the spectral graph filters in canonical GNN models, e.g., GCN/SGC~\cite{kipf2017semi,Wu2019}, APPNP~\cite{Klicpera2018}, and GDC~\cite{gasteiger2019diffusion}, can be represented by the following forms:
\begin{small}
\begin{equation}
\begin{gathered}
\textstyle \ZM = \hat{\AM}^K \XM,\ \ZM = \hat{\AM}^K \XM,\ \text{and}\ \ZM = \sum_{k=0}^K {\alpha^k}\hat{\AM}^k \XM,
\end{gathered}
\end{equation}
\end{small}
where $\alpha$ is a weight coefficient. 
Their corresponding spectral graph filters $g_\theta(\hat{\LM})$ are basically link prediction heuristics: $k$-hop random walk, PPR, and HKPR matrices in Section~\ref{sec:linpred}, whose polynomial coefficients $\boldsymbol{\theta}_k=0$ $(0\le k\le K-1)$ and $\boldsymbol{\theta}_K=1$ for GCN/SGC, and $\boldsymbol{\theta}_k$ is ${\alpha^k}$ and $\frac{e^{-\alpha}\alpha^k}{k!}$ for APPNP and GDC, respectively.
Accordingly, the spectral filtering function $g_\theta(\IM-\boldsymbol{\Lambda})$ is $f(\boldsymbol{\Lambda})$ in Table~\ref{tab:linkpred}.

\section{Methodology}
This section presents our \algo model for link sign prediction in SBGs. More concretely, \algo includes $K$ layers of sign-aware spectral convolutional operations that are built on our Gegenbauer polynomial-based spectral graph filters (Sections~\ref{subsec:design} and \ref{sec:layer}), our spectral decomposition approaches for initializing node features by extracting inter- and intra-partition patterns from SBGs (Section~\ref{sec:initial}), as well as the model training objective in Section~\ref{sec:objective}.

\begin{figure*}[!t]
    \centering
    \includegraphics[width=0.9\textwidth]{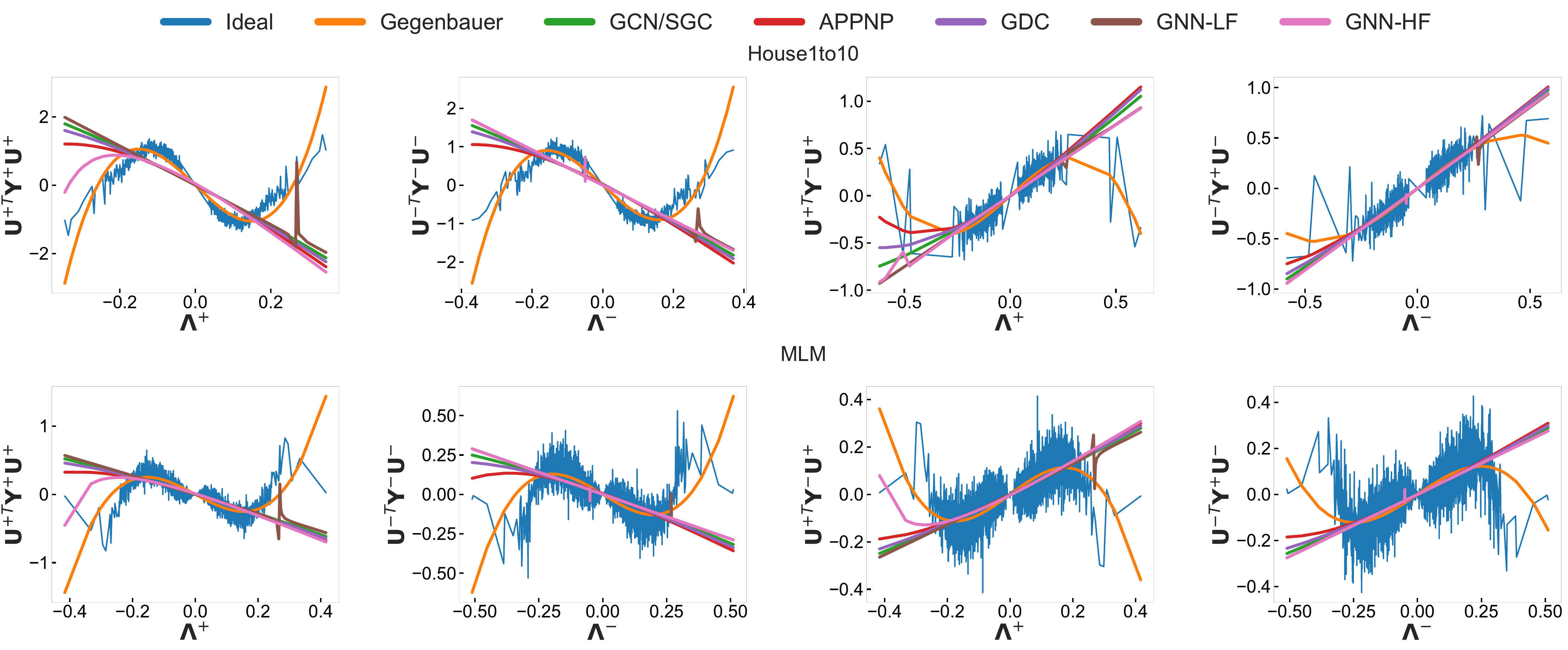}
    \vspace{-3ex}
    \caption{The relations between ${\Lambda}^{+}$ and ${{\UM}^{+}}^{\top}\YM^{+}{\UM}^{+}$,  ${\boldsymbol{\Lambda}}^{-}$ and ${{\UM}^{-}}^{\top}\YM^{-}{\UM}^{-}$, ${\boldsymbol{\Lambda}}^{+}$ and ${{\UM}^{+}}^{\top}\YM^{-}{\UM}^{+}$, ${\boldsymbol{\Lambda}}^{-}$ and ${{\UM}^{-}}^{\top}\YM^{+}{\UM}^{-}$.}
    \label{fig:spectral_analysis}
\end{figure*}

\begin{table}[!t]
\centering
\caption{Spectral filters and hyperparameters used in Fig.~\ref{fig:spectral_analysis}.}
\label{tab:spectral_summary}
\vspace{-3mm}
\resizebox{\columnwidth}{!}{%
\begin{tabular}{lccc}
    \hline
    \text{Model} & \text{Heuristic} & \( g_\theta(\boldsymbol{\Lambda})\) & \text{Hyperparameters} \\
    \hline
    \text{GCN/SGC~\cite{kipf2017semi,Wu2019}} & $K$-hop RW & \(\boldsymbol{\Lambda}^K\) & \(K=3\) \\
    \text{APPNP \cite{Klicpera2018}} & PPR & \(\sum_{k=0}^K {\alpha^k}\boldsymbol{\Lambda}^k\) & \(\alpha=0.9, K=7\) \\
    \text{GDC~\cite{gasteiger2019diffusion}} & HKPR & \(\sum_{k=0}^K \frac{e^{-\alpha}\alpha^k}{k!}\boldsymbol{\Lambda}^k\) & \(\alpha=2, K=7\) \\
    \text{GNN-LF \cite{Zhu2021}} & - & \(\frac{1-(1-\beta)(1-\boldsymbol{\Lambda})}{1-\left(2-\beta+\frac{1}{\alpha}\right)(1-\boldsymbol{\Lambda})}\) & \(\alpha=0.1, \beta=0.75\) \\
    \text{GNN-HF \cite{Zhu2021}} & - & \(\frac{1+\beta(1-\boldsymbol{\Lambda})}{1-\left(1-\beta-\frac{1}{\alpha}\right)(1-\boldsymbol{\Lambda})}\) & \(\alpha=0.1, \beta=1.0\) \\
    \text{Ours} & - & Eq.~\eqref{eq:jacobi} & \(\alpha=1.5,k=3\) \\
    \hline
\end{tabular}
}
\end{table}

\subsection{Design of Spectral Graph Filters}\label{subsec:design}
Instead of employing classic spectral graph filters that are merely designed for positive links, this section investigates the design of the filter $g_{\boldsymbol{\theta}}(\boldsymbol{\Lambda})$ for SBGs on the basis of theoretical analyses and empirical observations pertaining to the relations between observed and future positive/negative edges, respectively.

\subsubsection{\bf Spectral Transformation of Link Sign Prediction}
Let $\YM^{+}$ and $\YM^{-}$ contain the missing positive and negative links in the existing adjacency matrices $\hat{\AM}^{+}$ and $\hat{\AM}^{-}$, respectively. The link sign prediction task can be framed as looking for functions that map known links to the missing links with minimal error. 
Since there are positive and negative links in both training sets $\hat{\AM}^{+}$, $\hat{\AM}^{-}$ and test sets $\YM^{+}$, $\YM^{-}$, there are four possible functions, i.e., $f^{(++)}(\cdot)$, $f^{(--)}(\cdot)$, $f^{(-+)}(\cdot)$, and $f^{(+-)}(\cdot)$. 
In mathematical terms, the goal is to optimize
\begin{small}
\begin{equation}\label{eq:pos-pos-neg-pos}
\begin{aligned}
\min_{f^{(++)}\in \mathcal{F}}{\|f^{(++)}(\hat{\AM}^{+})-\YM^{+}\|_F},\ 
\min_{f^{(-+)}\in \mathcal{F}}{\|f^{(-+)}(\hat{\AM}^{-})-\YM^{+}\|_F}
\end{aligned}
\end{equation}
\begin{equation}\label{eq:neg-neg-pos-neg}
\begin{aligned}
\min_{f^{(--)}\in \mathcal{F}}{\|f^{(--)}(\hat{\AM}^{-})-\YM^{-}\|_F}
,\ \min_{f^{(-+)}\in \mathcal{F}}{\|f^{(+-)}(\hat{\AM}^{+})-\YM^{-}\|_F}.
\end{aligned}
\end{equation}
\end{small}
Intuitively, $f^{(++)}$ infers positive links from known positive edges using the common heuristic in social networks, namely, 
{\em the friend of my friend is my friend}~\cite{granovetter1973strength}, while $f^{(-+)}$ seeks to derive positive links from known negative edges by following ``{\em the enemy of my enemy is my friend}'' in social psychology~\cite{cartwright1956structural}.

\begin{lemma}\label{lem:curve-fitting}
Let $\hat{\AM}^{+}={\UM}^{+}{\boldsymbol{\Lambda}}^{+}{{\UM}^{+}}^{\top}$ be the eigendecomposition of $\hat{\AM}^{+}$. Then, the problem $\min_{f^{(++)}\in \mathcal{F}}{\|f^{(++)}(\hat{\AM}^{+})-\YM^{+}\|_F}$ is equivalent to $\min_{f^{(++)}\in \mathcal{F}}{\sum_{i=1}^{|\U|+|\V|}{(f^{(++)}({\boldsymbol{\Lambda}_{i,i}}^{+})-{{\UM}_{\cdot,i}^{+}}^{\top}\YM^{+}{\UM}_{\cdot,i}^{+})^2}}$.
\end{lemma}

Analogously, if $\hat{\AM}^{-}={\UM}^{-}{\boldsymbol{\Lambda}}^{-}{{\UM}^{-}}^{\top}$ is the eigendecomposition of $\hat{\AM}^{-}$, we have
\begin{small}
\begin{equation*}
\begin{aligned}
\min_{f^{(-+)}\in \mathcal{F}}{\|f^{(-+)}(\hat{\AM}^{-})-\YM^{+}\|_F} \Leftrightarrow 
{\sum_{i=1}^{|\U|+|\V|}{(f^{(-+)}({\boldsymbol{\Lambda}_{i,i}}^{-})-{{\UM}_{\cdot,i}^{-}}^{\top}\YM^{+}{\UM}_{\cdot,i}^{-})^2}},
\\
\min_{f^{(--)}\in \mathcal{F}}{\|f^{(--)}(\hat{\AM}^{-})-\YM^{-}\|_F} \Leftrightarrow 
{\sum_{i=1}^{|\U|+|\V|}{(f^{(--)}({\boldsymbol{\Lambda}_{i,i}}^{-})-{{\UM}_{\cdot,i}^{-}}^{\top}\YM^{-}{\UM}_{\cdot,i}^{-})^2}},
\\
\min_{f^{(+-)}\in \mathcal{F}}{\|f^{(+-)}(\hat{\AM}^{+})-\YM^{-}\|_F} \Leftrightarrow 
{\sum_{i=1}^{|\U|+|\V|}{(f^{(+-)}({\boldsymbol{\Lambda}_{i,i}}^{+})-{{\UM}_{\cdot,i}^{+}}^{\top}\YM^{-}{\UM}_{\cdot,i}^{+})^2}}.
\end{aligned}
\end{equation*}
\end{small}

\subsubsection{\bf Curve Fitting with Spectral Filtering Functions}
Based on the above spectral transformations, the choice of transformation functions $f^{(++)}(\cdot)$, $f^{(--)}(\cdot)$, $f^{(-+)}(\cdot)$, and $f^{(+-)}(\cdot)$ for link sign prediction can be reduced to a {\em least-squares curve fitting problem}. 
Take $f^{(++)}(\cdot)$ as an example.
Intuitively, when we let eigenvalues ${\boldsymbol{\Lambda}_{i,i}}^{+}$ ($1\le i\le |\U|+|\V|$) be the $x$-axis, the curve of an ideal $f^{(++)}({\boldsymbol{\Lambda}_{i,i}}^{+})$ should fit the curve of ${{\UM}_{\cdot,i}^{+}}^{\top}\YM^{+}{\UM}_{\cdot,i}^{+}$.
Doing so facilitates us to find spectral filters that are best suited to the link sign prediction by empirical observations on real SBGs.

Inspired by this, on the {\em MLM} and {\em House1to10} datasets (Table~\ref{tbl:datasets}), we empirically study the choices of $f^{(++)}(\cdot)$, $f^{(--)}(\cdot)$, $f^{(-+)}(\cdot)$, and $f^{(+-)}(\cdot)$ using the five spectral filtering functions $g_{\boldsymbol{\theta}}(\cdot)$ adopted in GCN/SGC, APPNP, GDC, GNN-LF, and GNN-HF with hyperparameters suggested in their respective papers (see Table~\ref{tab:spectral_summary}), respectively. 
In particular, we refer to the regions where eigenvalues with small (near $-1.0$), middle (roughly within the interval $[-0.2,0.2]$), and large values (near $1.0$), as high-, mid-, and low-frequency areas, respectively, and their corresponding ideal values as high-, mid-, and low-frequency signals, respectively~\cite{shuman2013emerging}.
According to Fig.~\ref{fig:spectral_analysis}, we can make the following observations.
\begin{itemize}[leftmargin=*]
\item ${{\UM}_{\cdot,i}^{+}}^{\top}\YM^{+}{\UM}_{\cdot,i}^{+}$ and ${{\UM}_{\cdot,i}^{-}}^{\top}\YM^{-}{\UM}_{\cdot,i}^{-}$ are negatively correlated to the eigenvalues in ${\boldsymbol{\Lambda}}^{+}$ and ${\boldsymbol{\Lambda}}^{-}$, respectively, when they are in the mid-frequency area, but are positively correlated to the eigenvalues in the high- and low-frequency areas.
\item In comparison, ${{\UM}_{\cdot,i}^{+}}^{\top}\YM^{-}{\UM}_{\cdot,i}^{+}$ (resp. ${{\UM}_{\cdot,i}^{-}}^{\top}\YM^{+}{\UM}_{\cdot,i}^{-}$) exhibits completely opposite behaviors to the eigenvalues in ${\boldsymbol{\Lambda}}^{+}$ (resp. ${\boldsymbol{\Lambda}}^{-}$).
\item The curves by the spectral filtering functions closely approximate the ideal curves in the mid-frequency area, but deviate significantly from them with opposite trends outside this region.
\end{itemize}

The above phenomenon indicates that existing spectral filters $g_{\boldsymbol{\theta}}(\cdot)$ are incompetent for capturing low- and high-frequency signals in both positive and negative links.
Their limited expressive power can be ascribed to the fact that they are composed of a series of monomial bases that are not orthogonal to each other w.r.t. weight functions~\cite{Wang2022a}.

\subsubsection{\bf Gegenbauer Polynomial Filters}
In response, we propose to employ the complex and orthogonal polynomial, i.e., the Gegenbauer polynomial function, as the transformation functions $f^{(++)}(\cdot)$, $f^{(--)}(\cdot)$, $f^{(-+)}(\cdot)$, and $f^{(+-)}(\cdot)$. Specifically, given a diagonal matrix $\boldsymbol{\Lambda}$ containing eigenvalues, its Gegenbauer polynomial bases are defined in a recursive form: 
\begin{small}
\begin{equation}\label{eq:jacobi}
\begin{gathered}
J^{\alpha}_0(\boldsymbol{\Lambda})=\IM,\ J^{\alpha}_1(\boldsymbol{\Lambda})=\left(\alpha+\frac{1}{2}\right)\cdot\boldsymbol{\Lambda},\\
\text{$k \geq 2$},\ J^{\alpha}_k(\boldsymbol{\Lambda})=\omega_k \boldsymbol{\Lambda} \cdot J^{\alpha}_{k-1}(\boldsymbol{\Lambda})-\omega_k^{\prime}\cdot J^{\alpha}_{k-2}(\boldsymbol{\Lambda}),
\end{gathered}
\end{equation}
\end{small}
where the coefficient $\alpha \ge -\frac{1}{2}$ and weights satisfy
\begin{small}
\begin{equation*}
\begin{gathered}
\omega_k =\frac{(2k+2\alpha-1)(k+\alpha-1)}{ k(k+2\alpha-1)}\ \text{and}\ \omega_k^{\prime} =\frac{(k+\alpha-\frac{1}{2})(k+\alpha-\frac{3}{2})}{k(k+2\alpha-1)}.
\end{gathered}
\end{equation*}
\end{small}

\begin{theorem}\label{thm:gegenbauer}
For $K\ge \mathbb{N}_0$ and $\alpha, a\ge -\frac{1}{2}$, the Gegenbauer polynomial basis $J^{\alpha}_K(\boldsymbol{\Lambda})$ is a $K$-order Gegenbauer polynomial, i.e., \(J^{\alpha}_K(\boldsymbol{\Lambda})=\sum_{k=0}^K{c_k^K\cdot J^{a}_k(\boldsymbol{\Lambda})}\), where \(c_k^K\) is the polynomial coefficient.
\end{theorem}

Theorem~\ref{thm:gegenbauer} indicates that the $k$-th Gegenbauer polynomial basis \(J^{\alpha}_k(\boldsymbol{\Lambda})\) is equivalent to a $k$-order Gegenbauer polynomial, which implies that we can directly adopt \(J^{\alpha}_k(\boldsymbol{\Lambda})\) as the spectral filtering function, and hence, the transformation functions $f^{(++)}(\cdot)$, $f^{(--)}(\cdot)$, $f^{(-+)}(\cdot)$, and $f^{(+-)}(\cdot)$.
Accordingly, the corresponding spectral graph filters $g_{\boldsymbol{\theta}}(\cdot)$ for positive and negative links are defined as in the following recursive form\footnote{The closed‐form expression can be found in Appendix~\ref{eq:closed-form}.}
\begin{small}
\begin{equation}
\begin{gathered}
J^{\alpha}_0({\hat{\AM}}^{+})=\IM,\ J^{\alpha}_1(\hat{\AM}^{+})=({\alpha+1})\cdot\hat{\AM}^{+},\\
J^{\alpha}_k(\hat{\AM}^{+})=\omega_k \hat{\AM}^{+}\cdot J^{\alpha}_{k-1}(\hat{\AM}^{+})-\omega_k^{\prime}\cdot J^{\alpha}_{k-2}(\hat{\AM}^{+})\ \forall{k \geq 2},\ \text{and}
\end{gathered}
\end{equation}
\end{small}
\begin{small}
\begin{equation}
\begin{gathered}
J^{\alpha}_0({\hat{\AM}}^{-})=\IM,\ J^{\alpha}_1(\hat{\AM}^{-})=({\alpha+1})\cdot\hat{\AM}^{-},\\
J^{\alpha}_k(\hat{\AM}^{-})=\omega_k \hat{\AM}^{-}\cdot J^{\alpha}_{k-1}(\hat{\AM}^{-})-\omega_k^{\prime}\cdot J^{\alpha}_{k-2}(\hat{\AM}^{-})\ \forall{k \geq 2}.
\end{gathered}
\end{equation}
\end{small}

As plotted in Fig.~\ref{fig:spectral_analysis}, we can observe that the Gegenbauer polynomial basis $J^{\alpha}_k(\cdot)$ well fits the target curves in the four cases
when $\alpha=1.0$ and $k=3$, especially in the low- and high-frequency areas, demonstrating its high efficacy in link sign prediction on SBGs. 

\eat{
\begin{lemma}
\begin{proof}
Let $\mathbf{x}\in \mathbb{R}^d$.

$\mathbf{z} = \sum_{k=0}^K{\boldsymbol{\theta}_k J^{\alpha}_k(\hat{\AM}) \mathbf{x}}$

Let $x_i=\UM_{\cdot,i}^\top\mathbf{x}$
\begin{align*}
\frac{\partial \frac{1}{2}\|\mathbf{z}-\mathbf{z}^{\star}\|_F^2}{\boldsymbol{\theta}_k\boldsymbol{\theta}_\ell} & = \mathbf{x}^\top J^{\alpha}_k(\hat{\AM}) J^{\alpha}_\ell(\hat{\AM}) \mathbf{x} \\
& = \mathbf{x}^\top \UM J^{\alpha}_k(\boldsymbol{\Lambda}) \UM^\top \UM J^{\alpha}_\ell(\boldsymbol{\Lambda}) \UM^\top \mathbf{x} \\
& = \sum_{i=1}^{|\U|+|\V|}{J^{\alpha}_k(\lambda_i) J^{\alpha}_\ell(\lambda_i) x_i^2}
\end{align*}

Let $w(\lambda)=(1-\lambda^2)^{\alpha-1/2}$ be the weight function. For $k\neq \ell$ and $\alpha > -\frac{1}{2}$,
 \begin{align*}
h(k,\ell)= \int_{\lambda=-1}^{1} J^{\alpha}_k(\lambda) J^{\alpha}_{\ell}(\lambda) w(\lambda) \,d\lambda = 0
 \end{align*}

$J^{\alpha}_k(\lambda)$ are orthogonal polynomials on the interval $[-1,1]$.
\end{proof}
\end{lemma}
}


\stitle{Remark}
Gegenbauer polynomials generalize other well-known polynomials, i.e.,
\begin{itemize}[leftmargin=*]
\item When $\alpha=0$ or $1.0$, $J^{\alpha}_k(\boldsymbol{\Lambda})$ is Chebyshev polynomials;
\item When $\alpha=0.5$, $J^{\alpha}_k(\boldsymbol{\Lambda})$ is Legendre polynomials.
\end{itemize}

\subsection{Sign-aware Spectral Convolutional Layers}\label{sec:layer}

We denote by $\HM^{(\ell)}$ ($0\le \ell\le L-1$) the output node embeddings at the $\ell$-th layer of the \algo model. Initially, 
\begin{equation*}
\HM^{(0)} = \XM\WM^{(0)},
\end{equation*}
where $\WM^{(0)}$ is learnable parameters and $\XM$ denotes the spectral node features extracted from the input graph topology $\PAM$ and $\NAM$. The construction of $\XM$ is deferred to the succeeding section.
As pinpointed in the preceding section, both the positive and negative links in $\G$ are conducive to the prediction of unknown links in $\YM^{+}$ and $\YM^{-}$, respectively. As such, at the $(\ell+1)$-th layer, \algo creates $\HM^{(\ell+1)}_{\text{pos}}$ and $\HM^{(\ell+1)}_{\text{neg}}$ by applying our Gegenbauer polynomial filter $J^{\alpha}_{\ell+1}$ on $\HM^{(\ell)}$ with $\hat{\AM}^{+}$ and $\hat{\AM}^{-}$, respectively, as follows:
\begin{small}
\begin{equation}
\begin{aligned}
\HM^{(\ell+1)}_{\text{pos}} &= \textsf{PReLU}(\delta\cdot J^{\alpha}_{\ell+1}(\hat{\AM}^{+})\HM^{(\ell)}\WM^{(\ell+1)}_{\text{pos}})\\
\HM^{(\ell+1)}_{\text{neg}} &= \textsf{PReLU}(\delta\cdot J^{\alpha}_{\ell+1}(\hat{\AM}^{-})\HM^{(\ell)}\WM^{(\ell+1)}_{\text{neg}}),
\end{aligned}
\end{equation}
\end{small}
where $\WM^{(\ell+1)}_{\text{pos}}$ and $\WM^{(\ell+1)}_{\text{neg}}$ are layer-specific
learnable weights, $\delta$ is a weight hyperparameter, and $\textsf{PReLU}(\cdot)$ stands for the parametric ReLU, i.e., the non-linear activation function. Intuitively, $\HM^{(\ell+1)}_{\text{pos}}$ (resp. $\HM^{(\ell+1)}_{\text{neg}}$) encodes both structural connectivity and positive (resp. negative) relationships from $\hat{\AM}^{+}$ (resp. $\hat{\AM}^{-}$). Apart from them, we further apply a non-linear transformation on $\HM^{(\ell)}$ to get 
\begin{small}
\begin{equation}
\HM^{(\ell+1)}_{\text{org}} = \textsf{PReLU}(\HM^{(\ell)}\WM^{(\ell+1)}_{\text{org}}),
\end{equation}
\end{small}
which carries the accumulated information in the past.

In turn, as in Eq.~\eqref{eq:Hk1}, the node embeddings $\HM^{(\ell+1)}$ output at the $(\ell+1)$-th model layer is obtained by fusing $\HM^{(\ell+1)}_{\text{org}}$, $\HM^{(\ell+1)}_{\text{pos}}$ and $\HM^{(\ell+1)}_{\text{neg}}$ via a concatenation, followed by a linear transformation with learnable weights $\WM^{(\ell+1)}_{\text{cat}}$.
\begin{small}
\begin{equation}\label{eq:Hk1}
\HM^{(\ell+1)} = \left(\HM^{(\ell+1)}_{\text{org}} \mathbin\Vert \HM^{(\ell+1)}_{\text{pos}} \mathbin\Vert \HM^{(\ell+1)}_{\text{neg}}\right)\cdot\WM^{(\ell+1)}_{\text{cat}}
\end{equation}
\end{small}
In doing so, \algo is able to inject sign-aware structural information and their sophisticated fusion and interactions into the node embeddings.
The output $\HM^{(\ell)}$ at the $L$-th layer is used as the final node representations $\ZM$ for nodes in $\U$ and $\V$.

\stitle{Linearization Analysis} Next, we unveil the node representations $\ZM$ learned by the model after removing non-linearity from the above $L$ convolutional layers. First, notice that the concatenation and linear transformation in Eq.~\eqref{eq:Hk1} can be reformulated as a summation as follows:
\begin{small}
\begin{align*}
\HM^{(\ell+1)} = & J^{\alpha}_{\ell+1}(\hat{\AM}^+) \HM^{(\ell)} \WM^{(\ell+1)}_{\text{pos}}\WM^{(\ell+1)}_{\text{cat},\text{pos}} + J^{\alpha}_{\ell+1}(\hat{\AM}^-) \HM^{(\ell)} \WM^{(\ell+1)}_{\text{neg}} \WM^{(\ell+1)}_{\text{cat},\text{neg}}\\
& + \HM^{(\ell)} \WM^{(\ell+1)}_{\text{org}} \WM^{(\ell+1)}_{\text{cat},\text{org}}
\end{align*}
\end{small}
since the weight matrix $\WM^{(\ell+1)}_{\text{cat}}$ is composed of three matrix blocks, i.e., ${\WM^{(\ell+1)}_{\text{cat}}}^\top={\WM^{(\ell+1)}_{\text{cat},\text{pos}}}^\top \mathbin\Vert {\WM^{(\ell+1)}_{\text{cat},\text{neg}}}^\top \mathbin\Vert {\WM^{(\ell+1)}_{\text{cat},\text{org}}}^\top$. Given that
\begin{small}
\begin{align*}
\HM^{(1)} = & J^{\alpha}_{1}(\hat{\AM}^+) \XM\WM^{(0)} \WM^{(1)}_{\text{pos}}\WM^{(1)}_{\text{cat},\text{pos}} + J^{\alpha}_{1}(\hat{\AM}^-) \XM\WM^{(0)} \WM^{(1)}_{\text{neg}} \WM^{(1)}_{\text{cat},\text{neg}}\\
& + \XM\WM^{(0)} \WM^{(1)}_{\text{org}} \WM^{(1)}_{\text{cat},\text{org}}
\end{align*}
\end{small}
and $\mathcal{P}^{(\ell)}=\{J^{\alpha}_\ell(\hat{\AM^+}), J^{\alpha}_\ell(\hat{\AM^-}), \IM\},\ 1\le \ell\le L$, we can derive that
\begin{small}
\begin{equation*}
\ZM = \HM^{(L)} = \sum_{\PM^{(1)}\in \mathcal{P}^{(1)}} \cdots\sum_{\PM^{(L)}\in \mathcal{P}^{(L)}}{{\left(\prod_{\ell=1}^L{\PM^{(\ell)}}\right)}\XM\cdot W(\PM^{(1)},\ldots,\PM^{(L)})},
\end{equation*}
\end{small}
where the weight matrix $W(\PM^{(1)},\ldots,\PM^{(L)})=\WM^{(0)}\WM^{(1)}\cdots \WM^{(L)}$ for each term is a product of weight matrices at each layer and $\ 1\le \ell\le L$,
\begin{small}
\begin{equation*}
\WM^{(\ell)} = 
\begin{cases}
\WM^{(\ell)}_{\text{pos}}\WM^{(\ell)}_{\text{cat},\text{pos}} & \text{if $\PM^{(\ell)}=J^{\alpha}_\ell(\hat{\AM^+})$,}\\
 \WM^{(\ell)}_{\text{neg}}\WM^{(\ell)}_{\text{cat},\text{neg}} & \text{if $\PM^{(\ell)}=J^{\alpha}_\ell(\hat{\AM^-})$,}\\
 \WM^{(\ell)}_{\text{org}} \WM^{(\ell)}_{\text{cat},\text{org}} & \text{otherwise}.
\end{cases}
\end{equation*}
\end{small}

Particularly, $\ZM$ can be decomposed into a sum of $3^K$ different terms, each of which applies $K$ operators (spectral filters $J^{\alpha}_\ell(\hat{\AM}^+)$, $J^{\alpha}_\ell(\hat{\AM}^-)$ or identity mapping $\IM$) $\PM^{(1)}\cdots\PM^{(L)}$ over initial node features $\XM$ followed by a linear transformation with unique weights $W(\PM^{(1)},\ldots,\PM^{(L)})$. Such a sophisticated combination empowers \algo to comprehensively and accurately extract and encode the sign-aware topological features in SBGs.

\subsection{Spectral Feature Initialization}\label{sec:initial}
In lieu of learning node representations $\ZM$ from randomly initialized node features $\XM$, \algo resorts to extracting structural semantics underlying the input topology of SBG $\G$ as the initial node feature matrix $\XM\in \mathbb{R}^{(|\U|+|\V|)\times d}$. Recall that the nodes in $\G$ are partitioned into $\U$ and $\V$ by their heterogeneous sources. Intuitively, $\XM$ should capture not only the {\em inter-partition relations}, i.e., proximity of nodes across $\U$ and $\V$,
but also the {\em intra-partition relations}, i.e., similarities of nodes within the same partition. For ease of notation, we use $\UM$ (resp. $\VM$) to represent the feature vectors of nodes in $\U$ (resp. $\V$). Accordingly, 
$\footnotesize \textstyle \XM=\left[ 
\begin{array}{c} 
  \UM  \\
  \VM 
\end{array} 
\right] $.

\stitle{Inter-Partition Relation Preservation}
First, we formulate our objective function for preserving the inter-partition relations into $\UM$ and $\VM$ as follows:
\begin{small}
\begin{equation}
\min_{\substack{\UM\in \mathbb{R}^{|\U|\times d}\\ \VM \in \mathbb{R}^{|\V|\times d}}}\sum_{(u_i,v_j)\in \PEDG}{\|\UM_{u_i}-\VM_{v_j}\|^2_2} - \sum_{(u_i,v_j)\in \NEDG}{\|\UM_{u_i}-\VM_{v_j}\|^2_2},
\end{equation}
\end{small}
which is to minimize the embedding distance of nodes in $\U$ and $\V$ with positive links and maximize their distance with negative links. 
Let $\LM$ be the Laplacian matrix of $\G$, i.e., $\LM = \DM - \tilde{\AM}$, where $\DM$ stands for a $(|\U|+|\V|)^2$ diagonal matrix whose each diagonal element $\DM_{x,x} = \sum_{y\in \U\cup \V}{\tilde{\AM}_{x,y}}$.
Based thereon, we can rewrite the above objective in the following equivalent form:
\begin{small}
\begin{equation}\label{eq:obj1}
\min_{\XM\in \mathbb{R}^{(|\U|+|\V|)\times d}}\sum_{u_i\in |\U|,v_j\in |\V|}{\tilde{\AM}_{u_i,v_j}\cdot \|\XM_{u_i}-\XM_{v_j}\|^2_2} = \Tr(\XM^{\top}\LM\XM).
\end{equation}
\end{small}

\stitle{Intra-Partition Relation Preservation}
Next, we encode the inter-partition relations of nodes into the feature vectors $\XM$. We define the intra-partition similarity of any node pair $u_i,u_j\in \U$ by
\begin{small}
\begin{equation}
\begin{gathered}
\textstyle s(u_i,u_j) = \frac{\PAM_{u_i}\cdot \PAM_{u_j} + \NAM_{u_i}\cdot \NAM_{u_j}}{\sqrt{|\PNGH|_{u_i}+|\NNGH|_{u_i}} \sqrt{|\PNGH|_{u_j}+|\NNGH|_{u_j}}},
\end{gathered}
\end{equation}
\end{small}
where $\PAM_{u_i}\cdot \PAM_{u_j}$ + $\NAM_{u_i}\cdot \NAM_{u_j}$ denotes the total amount of nodes in $\V$ that directly connect to both $u_i$ and $u_j$, and the denominator signifies a normalization. To explain, if we regard $\U$ as users, $\V$ as movies, and $\PEDG$/$\NEDG$ as likes and dislikes of users to movies, $s(u_i,u_j)$ quantifies the similarity of the preferences of $u_i$ and $u_j$ on movies.
In the same vein, we can define the intra-partition similarity $s(v_i,v_j)$ of any node pair $v_i,v_j\in \V$.

We then formulate our second objective function for capturing the inter-partition similarity of homogeneous nodes in $\UM$ and $\VM$ as
\begin{small}
\begin{equation*}
\begin{split}
\min_{\substack{\UM\in \mathbb{R}^{|\U|\times d}\\ \VM \in \mathbb{R}^{|\V|\times d}}}\sum_{u_i,u_j\in \U}{s(u_i,u_j)\cdot\|\UM_i-\UM_j\|^2_2} +\sum_{v_i,v_j\in \V}{s(v_i,v_j)\cdot\|\VM_i-\VM_j\|^2_2}\\
- \left(\sum_{u_i,u_j\in \U}{{s(u_i,u_j)}\cdot \|\UM_i\|^2_2} + \sum_{v_i,v_j\in \V}{{s(v_i,v_j)}\cdot \|\VM_i\|^2_2}\right).
\end{split}
\end{equation*}
\end{small}
The former two terms seek to enforce homogeneous nodes with high similarities to be close in the feature space, whereas the latter ones are two regularization terms. 

Let ${\BM}^{(r)}$ and ${\BM}^{(c)}$ be matrices obtained by applying $L_2$ normalization to the rows and columns of the bi-adjacency matrix $\AM$, respectively. It is easy to verify that $s(u_i,u_j) = {\BM}^{(r)}_{u_i}\cdot \BM^{(r)}_{u_j}\ \text{and}\ s(v_i,v_j) = {\BM}^{(c)}_{v_i}\cdot \BM^{(c)}_{v_j}$. Accordingly, we define $\BM$ as $\footnotesize \begin{pmatrix}
  \begin{matrix}
\bigzero
  \end{matrix}
  & \rvline & {\BM}^{(r)} \\
\hline
  {\BM}^{(c)} & \rvline &
  \begin{matrix}
\bigzero
  \end{matrix}
\end{pmatrix}$. 
\begin{equation}\label{eq:obj2}
\max_{\XM\in \mathbb{R}^{(|\U|+|\V|)\times d}}{\Tr(\XM^{\top}\BM\BM^{\top}\XM)}
\end{equation}
Eq. \eqref{eq:obj2} states that the above objective function is equivalent to a trace maximization problem.

\begin{table}
\centering
\caption{Dataset Statistics}\label{tbl:datasets}
\vspace{-3mm}
\renewcommand{\arraystretch}{0.8}
\begin{small}
\begin{tabular}{l|c|c|c|c|c}
\hline
{\bf Dataset} & {\bf $|\U|$} & {\bf $|\V|$} & {\bf $|\EDG^{+}\cup \EDG^{-}|$} & {\bf $\frac{|\EDG^{+}|}{|\EDG^{+}\cup \EDG^{-}|}$} &  {\bf  $\frac{|\EDG^{-}|}{|\EDG^{+}\cup \EDG^{-}|}$} \\ \hline
Review & 182 & 304 & 1,170 & 0.3966 & 0.6034 \\
Senate & 145 & 1,056 & 27,083 & 0.5531 & 0.4469 \\
Bonanza & 7,919 & 1,973 & 36,543 & 0.9798 & 0.0202 \\
House1to10 & 515 & 1,281 & 114,378 & 0.5396 & 0.4604 \\
MLM & 6,040 & 3,706 & 1,000,209 & 0.5752 & 0.4248 \\
Amazon & 35,736 & 38,121 & 1,960,674 & 0.8058 & 0.1942 \\
 \hline
 \end{tabular}
\end{small}
\vspace{-2ex}
 \end{table}

\begin{table*}[!t]
\centering
\caption{Classification Performance (the higher the better).}
\label{tbl:classification-perf}
\vspace{-3ex}
\renewcommand{\arraystretch}{0.8}
\resizebox{\textwidth}{!}{
\begin{tabular}{|c|cc|cc|cc|cc|cc|cc|} 
\hline
\multirow{2}{*}{Method}                     & \multicolumn{2}{c|}{Amazon}                                                             & \multicolumn{2}{c|}{Bonanza}                                                                                & \multicolumn{2}{c|}{House1to10}                                                         & \multicolumn{2}{c|}{MLM}                                                                & \multicolumn{2}{c|}{Review}                                                                                                      & \multicolumn{2}{c|}{Senate}                                                                                                       \\ 
\cline{2-13}
                                            & AUC \textuparrow                                        & F1 \textuparrow                                         & AUC \textuparrow                                                            & F1 \textuparrow                                         & AUC \textuparrow                                        & F1 \textuparrow                                         & AUC \textuparrow                                        & F1 \textuparrow                                         & AUC \textuparrow                                                            & F1 \textuparrow                                                              & AUC \textuparrow                                                            & F1 \textuparrow                                                               \\ 
\hline
MSGNN \cite{he2022msgnn}                                      & 0.5381                                     & 0.4465                                     & 0.5623                                                         & 0.4946                                     & 0.4952                                     & 0.3176                                     & 0.5847                                     & 0.3653                                     & 0.4089                                                         & 0.3777                                                          & 0.4956                                                         & 0.3202                                                           \\
SDGNN \cite{huang2021sdgnn}                                      & 0.7562                                     & 0.5792                                     & \underline{0.6994}                                                 & 0.5541                                     & 0.6879                                     & 0.6305                                     & 0.7757                                     & 0.7033                                     & 0.7743                                                         & 0.7058                                                          & 0.7087                                                         & 0.6458                                                           \\
SGCN \cite{derr2018signed}                                       & 0.5667                                     & 0.4465                                     & 0.6630                                                         & 0.4946                                     & 0.6816                                     & 0.6192                                     & 0.7617                                     & 0.6907                                     & \underline{0.7768}                                                 & 0.6843                                                          & 0.7099                                                         & 0.6432                                                           \\
SiGAT \cite{huang2019signed}                                      & 0.7093                                     & 0.5106                                     & 0.6642                                                         & 0.4945                                     & 0.6764                                     & 0.6118                                     & 0.7612                                     & 0.6900                                     & 0.6433                                                         & 0.5986                                                          & 0.7008                                                         & 0.6422                                                           \\
SigMaNet \cite{10.1609/aaai.v37i6.25919}                                   & 0.4938                                     & 0.3753                                     & 0.4166                                                         & 0.1295                                     & 0.5072                                     & 0.4076                                     & 0.4542                                     & 0.4105                                     & 0.4893                                                         & 0.5011                                                          & 0.5030                                                         & 0.3501                                                           \\
SNEA \cite{li2020learning}                                       & 0.6855                                     & 0.5412                                     & 0.6799                                                         & 0.4946                                     & 0.6868                                     & 0.6253                                     & 0.7562                                     & 0.6864                                     & 0.7456                                                         & 0.6622                                                          & 0.7072                                                         & 0.6445                                                           \\
SSSNET \cite{he2022sssnet}                                     & 0.5066                                     & 0.4465                                     & 0.5626                                                         & 0.4946                                     & 0.5011                                     & 0.3141                                     & 0.6042                                     & 0.3676                                     & 0.4646                                                         & 0.3777                                                          & 0.5031                                                         & 0.3614                                                           \\
SBGCL \cite{zhang2023contrastive}                                      & -                                          & -                                          & 0.6739                                                         & 0.3187                                     & 0.7255                                     & 0.6289                                     & \multicolumn{1}{c}{-}                     & -                                          & 0.7235                                                         & 0.6696                                                          & 0.8497                                                         & 0.7542                                                           \\
SBGNN \cite{huang2021signed}                                      & -                                          & -                                          & 0.6781                                                         & \underline{0.5590}                             & \underline{0.9006}                             & \underline{0.8054}                             & -                                          & -                                          & 0.7549                                                         & \underline{0.7126}                                                  & \underline{0.8933}                                                 & \underline{0.8022}                                                   \\
SidNet \cite{jung2022signed}                                     & \underline{0.7754}                             & \underline{0.6126}                             & 0.6587                                                         & 0.5570                                     & 0.6900                                     & 0.6311                                     & \underline{0.7802}                             & \underline{0.7066}                             & 0.7175                                                         & 0.6597                                                          & 0.7047                                                         & 0.6387                                                           \\
SLGNN \cite{li2023signed}                                      & -                                          & -                                          & 0.5000                                                         & 0.4946                                     & 0.6132                                     & 0.6134                                     & 0.6827                                     & 0.6847                                     & 0.5554                                                         & 0.5558                                                          & 0.7874                                                         & 0.7876                                                           \\ 
\hline
\multicolumn{1}{|l|}{\algo ($\alpha=0$, Chebyshev)} & {\cellcolor[rgb]{0.776,0.776,0.776}}0.8017 & 0.6837                                     & 0.7258                                     & 0.5683                                     & {\cellcolor[rgb]{0.776,0.776,0.776}}0.9269 & {\cellcolor[rgb]{0.776,0.776,0.776}}0.8405 & 0.8087                                     & {\cellcolor[rgb]{0.776,0.776,0.776}}0.7289 & {\cellcolor[rgb]{0.776,0.776,0.776}}0.7952 & 0.7253                                     & 0.9042                                     & 0.8245                                      \\
 Improv.   &   +0.0263&+0.0711&+0.0264&+0.0093&+0.0263&+0.0351&+0.0285&+0.0223&+0.0184&+0.0127&+0.0109&+0.0223    \\ 
\multicolumn{1}{|l|}{\algo ($\alpha=0.5$, Legendre)}  & 0.8009                                     & 0.6833                                     & 0.7255                                     & 0.5666                                     & 0.9265                                     & 0.8384                                     & 0.8073                                     & 0.7258                                     & 0.7870                                     & 0.7184                                     & 0.9049                                     & {\cellcolor[rgb]{0.776,0.776,0.776}}0.8257  \\
 Improv.   &  +0.0255&+0.0707&+0.0261&+0.0076&+0.0259&+0.033&+0.0271&+0.0192&+0.0102&+0.0058&+0.0116&+0.0235  \\ 
\multicolumn{1}{|l|}{\algo ($\alpha=1.5$)}    & 0.8012                                     & {\cellcolor[rgb]{0.776,0.776,0.776}}0.6842 & {\cellcolor[rgb]{0.776,0.776,0.776}}0.7293 & {\cellcolor[rgb]{0.776,0.776,0.776}}0.5726 & 0.9260                                     & 0.8387                                     & {\cellcolor[rgb]{0.776,0.776,0.776}}0.8094 & 0.7276                                     & 0.7857                                     & {\cellcolor[rgb]{0.776,0.776,0.776}}0.7268 & {\cellcolor[rgb]{0.776,0.776,0.776}}0.9050 & 0.8250                                      \\
 Improv.   &  +0.0258&+0.0716&+0.0299&+0.0136&+0.0254&+0.0333&+0.0292&+0.021&+0.0089&+0.0142&+0.0117&+0.0228  \\ 
\hline
\end{tabular}
}
\end{table*}

\begin{table*}[!t]
\centering
\caption{Ablation Study.}
\label{tbl:ablation}
\vspace{-3ex}
\renewcommand{\arraystretch}{0.8}
\begin{small}
\resizebox{\textwidth}{!}{
\begin{tabular}{|c|cc|cc|cc|cc|cc|cc|} 
\hline
\multirow{2}{*}{Method}                     & \multicolumn{2}{c|}{Amazon}                                                             & \multicolumn{2}{c|}{Bonanza}                                                                                & \multicolumn{2}{c|}{House1to10}                                                         & \multicolumn{2}{c|}{MLM}                                                                & \multicolumn{2}{c|}{Review}                                                                                                      & \multicolumn{2}{c|}{Senate}                                                                                                       \\ 
\cline{2-13}
                                            & AUC \textuparrow                                        & F1 \textuparrow                                         & AUC \textuparrow                                                            & F1 \textuparrow                                         & AUC \textuparrow                                        & F1 \textuparrow                                         & AUC \textuparrow                                        & F1 \textuparrow                                         & AUC \textuparrow                                                            & F1 \textuparrow                                                              & AUC \textuparrow                                                            & F1 \textuparrow                                                               \\ 
\hline
\multicolumn{1}{|l|}{\algo ($\alpha=0$, Chebyshev)}      & \cellcolor[rgb]{0.776,0.776,0.776}0.8006 & 0.6834                                   & 0.6841                                   & \cellcolor[rgb]{0.776,0.776,0.776}0.5659 & 0.9236                                   & 0.8344                                   & \cellcolor[rgb]{0.776,0.776,0.776}0.8074 & 0.7250                                    & \cellcolor[rgb]{0.776,0.776,0.776}0.7795 & 0.6961                                   & 0.905                                    & \cellcolor[rgb]{0.776,0.776,0.776}0.8254 \\
\multicolumn{1}{|l|}{w/o negative ST}      & 0.7687                                   & 0.6643                                   & 0.6365                                   & 0.5046                                   & 0.9199                                   & 0.8262                                   & 0.8026                                   & 0.7176                                   & 0.7624                                   & 0.6827                                   & 0.9011                                   & 0.8180                                   \\
\multicolumn{1}{|l|}{w/o positive ST}       & 0.7683                                   & 0.6626                                   & 0.6712                                   & 0.5324                                   & 0.9203                                   & 0.8272                                   & 0.8006                                   & 0.7164                                   & 0.7247                                   & 0.6520                                   & 0.8937                                   & 0.8045                                   \\ 
\multicolumn{1}{|l|}{random embeddings w/o ST}       & 0.5132                                   & 0.4465                                   & 0.5595                                   & 0.4946                                   & 0.8593                                   & 0.7648                                   & 0.7324                                   & 0.6582                                   & 0.7149                                   & 0.6722                                   & 0.8371                                   & 0.7550                                   \\  \hline
\multicolumn{1}{|l|}{\algo ($\alpha=0.5$, Legendre)}     & 0.8004                                   & \cellcolor[rgb]{0.776,0.776,0.776}0.6838 & 0.7034                                   & 0.5500                                     & \cellcolor[rgb]{0.776,0.776,0.776}0.9243 & \cellcolor[rgb]{0.776,0.776,0.776}0.8369 & 0.8071                                   & 0.7243                                   & 0.7777                                   & \cellcolor[rgb]{0.776,0.776,0.776}0.7115 & \cellcolor[rgb]{0.776,0.776,0.776}0.9102 & 0.8249                                    \\
\multicolumn{1}{|l|}{w/o negative ST}      & 0.7673                                   & 0.6621                                   & 0.6452                                   & 0.5052                                   & 0.9163                                   & 0.8224                                   & 0.8013                                   & 0.7178                                   & 0.7051                                   & 0.6520                                   & 0.8988                                   & 0.8092                                    \\
\multicolumn{1}{|l|}{w/o positive ST}       & 0.7672                                   & 0.6621                                   & 0.6754                                   & 0.5165                                   & 0.9168                                   & 0.8244                                   & 0.7961                                   & 0.7188                                   & 0.7468                                   & 0.6799                                   & 0.8996                                   & 0.8133                                    \\
\multicolumn{1}{|l|}{random embeddings w/o ST}       & 0.5133                                   & 0.4465                                   & 0.5693                                   & 0.4946                                   & 0.8533                                   & 0.7656                                   & 0.7427                                   & 0.6735                                   & 0.7064                                   & 0.6691                                   & 0.8278                                   & 0.7464                                    \\  \hline
\multicolumn{1}{|l|}{\algo ($\alpha=1.5$)}      & 0.7996                                   & 0.6829                                   & \cellcolor[rgb]{0.776,0.776,0.776}0.7301 & 0.5619                                   & 0.9241                                   & 0.8324                                   & 0.8053                                   & \cellcolor[rgb]{0.776,0.776,0.776}0.7254 & 0.7529                                   & 0.6877                                   & 0.9048                                   & 0.8189                                    \\
\multicolumn{1}{|l|}{w/o negative ST}      & 0.7661                                   & 0.6610                                   & 0.6500                                   & 0.5054                                   & 0.9179                                   & 0.8262                                   & 0.8017                                   & 0.7210                                   & 0.7174                                   & 0.6722                                   & 0.8967                                   & 0.8035                                    \\
\multicolumn{1}{|l|}{w/o positive ST}       & 0.7666                                   & 0.6641                                   & 0.6729                                   & 0.5185                                   & 0.9169                                   & 0.8261                                   & 0.8014                                   & 0.7184                                   & 0.6182                                   & 0.6342                                   & 0.8997                                   & 0.8100                                    \\
\multicolumn{1}{|l|}{random embeddings w/o ST}       & 0.5132                                   & 0.4465                                   & 0.5697                                   & 0.4946                                   & 0.8679                                   & 0.7730                                   & 0.7299                                   & 0.6632                                   & 0.7205                                   & 0.6696                                   & 0.8483                                   & 0.7625                                    \\  \hline
\multicolumn{1}{|l|}{\algo (w/o ST)}      & 0.7864                                   & 0.6722                                   & 0.5928                                   & 0.4943                                   & 0.6886                                   & 0.6113                                   & 0.7871                                   & 0.7143                                   & 0.7557                                   & 0.7038                                   & 0.7003                                   & 0.6319                                    \\
\multicolumn{1}{|l|}{random embeddings w/o ST}      & 0.7672                                   & 0.6528                                   & 0.6426                                   & 0.5392                                   & 0.6749                                   & 0.5729                                   & 0.7397                                   & 0.6648                                   & 0.7097                                   & 0.6397                                   & 0.6847                                   & 0.6236                                \\  \hline
\end{tabular}
}
\end{small}
\end{table*}

\stitle{Spectral Decomposition} The joint optimization of Eq.~\eqref{eq:obj1} and Eq.~\eqref{eq:obj2} for the computation of $\XM$ is time-consuming due to the materialization of $\BM\BM^{\top}$ and numerous iterations needed for convergence. As a remedy, we additionally incorporate an orthogonal constraint on $\XM$ to facilitate the problem transformation and reduction. More precisely, we require the columns (i.e., feature dimensions) in $\XM$ to be orthogonal to each other. In doing so, the feature dimensions of $\XM$ are made decorrelate from each other, and thus, can embody richer structural information. 
\begin{theorem}[Ky Fan's trace minimization principle \cite{fan1949theorem}]\label{lem:ky}
Given a symmetric real matrix $\MM\in \mathbb{R}^{n\times n}$ with distinct eigenvalues $\lambda_1(\MM)$, $\lambda_2(\MM)$, $\cdots$, $\lambda_n(\MM)$, sorted by algebraic value in ascending order, eigenvectors $\boldsymbol{\Upsilon}$ and integer $d\le n$, we have
\begin{small}
\begin{equation}
\textstyle \min_{\XM^{\top}\XM=\IM_k}{\Tr(\XM^{\top}\MM\XM)} = {\Tr(\boldsymbol{\Upsilon}^{\top}\MM\boldsymbol{\Upsilon})} = \sum_{i=1}^{d}{\lambda_i(\MM)}.
\end{equation}
\end{small}
\end{theorem}
By Ky Fan's trace minimization principle in Theorem~\ref{lem:ky}, the optimal $\XM$ to Eq.~\eqref{eq:obj1} is the eigenvectors $\PhiM$ corresponding to the $d$-smallest eigenvalues of $\LM$. Analogously, Eq. \eqref{eq:obj2} is to find the $d$-largest eigenvectors $\PsiM$ of $\BM\BM^{\top}$, which are exactly the top-$d$ left singular vectors of $\BM$ according to Lemma~\ref{lem:svd-eig}.
\begin{lemma}\label{lem:svd-eig}
Let the columns of ${\PsiM}$ and $\boldsymbol{\alpha}$ be the $d$-largest eigenvectors of $\BM\BM^{\top}$ and the top-$d$ left singular vectors of $\BM$, respectively. Then, $\PsiM=\boldsymbol{\alpha}$.
\end{lemma}

As such, the resulting feature vectors $\XM$ of nodes are simply an amalgam of $\PhiM$ and $\PsiM$. Particularly, we combine the features from inter-partition and intra-partition relations using a weight $\mu$ as in
\begin{equation}
    \XM = \mu \cdot \PhiM + (1-\mu)\cdot \PsiM.
\end{equation}

\subsection{Model Training}\label{sec:objective}
After obtaining $\ZM$, we employ a 2-layer MLP to estimate
the sign score $y_{pred}$ between any two nodes $u_i\in \U$ and $v_j\in \V$:
\begin{equation}
y_{pred} = \textsf{sigmoid}(\textsf{MLP}(\ZM_{u_i} \mathbin\Vert \ZM_{v_j})).
\end{equation}
Particularly, the larger the $y_{pred}$ is, the higher the probability that the edge sign is positive. Contrarily, the smaller $y_{pred}$ is, the higher the probability that the edge sign is negative.
Following previous work \cite{huang2021signed,zhang2023contrastive}, we adopt the cross-entropy as the loss function for the link sign prediction task, which is formulated as
\begin{equation}
    \mathcal{L} = -y\cdot \log{(y_{pred})} + (1-y)\cdot \log{\left(1-y_{pred}\right)},
\end{equation}
where $y$ is the ground truth mapped from $\{-1,1\}$ to $\{0,1\}$.

\section{Experiments}
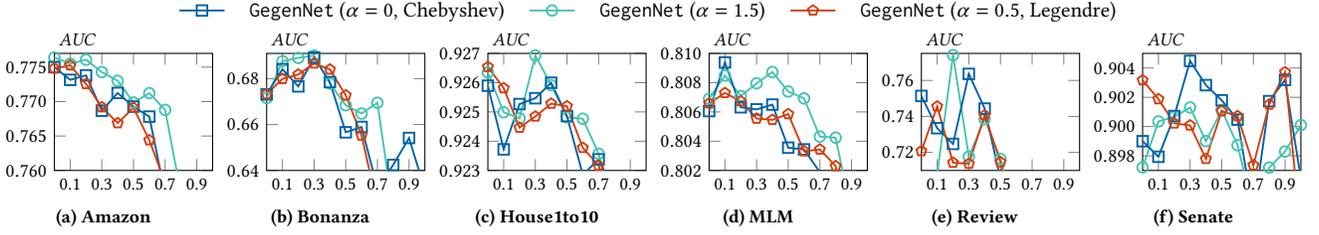
\begin{figure*}[!t]
\begin{small}
\begin{tikzpicture}
    \begin{customlegend}[legend columns=4,
        legend entries={{\algo{} (\(\alpha=0\), Chebyshev)},{\algo{} (\(\alpha=1.5\))}, {\algo{} (\(\alpha=0.5\), Legendre)}},
        legend style={at={(0.45,1.15)},anchor=north,draw=none,font=\small,column sep=0.25cm}]
    \addlegendimage{line width=0.25mm,mark size=2pt,mark=square, color=B2}
    \addlegendimage{line width=0.25mm,mark size=2pt,mark=o, color=B6}
    \addlegendimage{line width=0.25mm,mark size=2pt,mark=pentagon, color=O3}
    \end{customlegend}
\end{tikzpicture}
\\[-\lineskip]
\vspace{-4mm}
\hspace*{0mm}
\subfloat[{Amazon}]{
\begin{tikzpicture}[scale=1]\label{subfig:beta_Amazon}
    \begin{axis}[
        height=\columnwidth/2.7,
        width=\columnwidth/2.3,
        ylabel={\em AUC},
        xmin=0, xmax=10,
        ymin=0.76, ymax=0.777,
        xtick={1,3,5,7,9},
        xticklabel style = {font=\footnotesize},
        xticklabels={0.1,0.3,0.5,0.7,0.9},
        scaled y ticks = false,
        yticklabel style={/pgf/number format/fixed zerofill,/pgf/number format/precision=3},
        every axis y label/.style={at={(current axis.north west)},right=3mm,above=0mm},
        label style={font=\footnotesize},
        tick label style={font=\footnotesize},
    ]
        
    \addplot[line width=0.25mm,mark size=2pt,mark=o, color=B6] 
        plot coordinates {
(0, 0.776412368)
(1, 0.775498033)
(2, 0.776080489)
(3, 0.774285913)
(4, 0.77301091)
(5, 0.769871056)
(6, 0.771230102)
(7, 0.76879096)
(8, 0.756651223)
(9, 0.754784822)
(10, 0.590162992)
};
 \addplot[line width=0.25mm,mark size=2pt,mark=square, color=B2] 
        plot coordinates {
(0, 0.774998784)
(1, 0.773158193)
(2, 0.773812056)
(3, 0.768672347)
(4, 0.77125442)
(5, 0.76927042)
(6, 0.767819524)
(7, 0.756551862)
(8, 0.749333203)
(9, 0.744836688)
(10, 0.579551578)
};
    \addplot[line width=0.25mm,mark size=2pt,mark=pentagon, color=O3] 
        plot coordinates {
(0, 0.77482307)
(1, 0.775268912)
(2, 0.77256602)
(3, 0.769189477)
(4, 0.76687187)
(5, 0.769093633)
(6, 0.764394164)
(7, 0.758236289)
(8, 0.756797791)
(9, 0.747259736)
(10, 0.575931668)
};
    \end{axis}
\end{tikzpicture}\hspace{0mm}%
}%
\subfloat[{Bonanza}]{
\begin{tikzpicture}[scale=1]\label{subfig:beta_Bonanza}
    \begin{axis}[
        height=\columnwidth/2.7,
        width=\columnwidth/2.3,
        ylabel={\em AUC},
        xmin=0, xmax=10,
        ymin=0.64, ymax=0.691,
        xtick={1,3,5,7,9},
        xticklabel style = {font=\footnotesize},
        xticklabels={0.1,0.3,0.5,0.7,0.9},
        scaled y ticks = false,
        yticklabel style={/pgf/number format/fixed zerofill,/pgf/number format/precision=2},
        every axis y label/.style={at={(current axis.north west)},right=3mm,above=0mm},
        label style={font=\footnotesize},
        tick label style={font=\footnotesize},
    ]
        
    \addplot[line width=0.25mm,mark size=2pt,mark=o, color=B6] 
        plot coordinates {
(0, 0.671563983)
(1, 0.687603951)
(2, 0.689034402)
(3, 0.690057993)
(4, 0.677857697)
(5, 0.668401837)
(6, 0.664806008)
(7, 0.669507861)
(8, 0.622121155)
(9, 0.609336495)
(10, 0.632937551)
};
 \addplot[line width=0.25mm,mark size=2pt,mark=square, color=B2] 
        plot coordinates {
(0, 0.673168302)
(1, 0.684129953)
(2, 0.676635146)
(3, 0.689170599)
(4, 0.678497672)
(5, 0.656624675)
(6, 0.658954978)
(7, 0.632475019)
(8, 0.642411292)
(9, 0.654143691)
(10, 0.63161099)
};
    \addplot[line width=0.25mm,mark size=2pt,mark=pentagon, color=O3] 
        plot coordinates {
(0, 0.673274159)
(1, 0.679897666)
(2, 0.681824684)
(3, 0.686655641)
(4, 0.684034944)
(5, 0.672748804)
(6, 0.655240774)
(7, 0.624662995)
(8, 0.633921683)
(9, 0.625722408)
(10, 0.626787186)
};
    \end{axis}
\end{tikzpicture}\hspace{0mm}%
}%
\subfloat[{House1to10}]{
\begin{tikzpicture}[scale=1]\label{subfig:beta_House1to10}
    \begin{axis}[
        height=\columnwidth/2.7,
        width=\columnwidth/2.3,
        ylabel={\em AUC},
        xmin=0, xmax=10,
        ymin=0.923, ymax=0.927,
        xtick={1,3,5,7,9},
        xticklabel style = {font=\footnotesize},
        xticklabels={0.1,0.3,0.5,0.7,0.9},
        scaled y ticks = false,
        yticklabel style={/pgf/number format/fixed zerofill,/pgf/number format/precision=3},
        every axis y label/.style={at={(current axis.north west)},right=3mm,above=0mm},
        label style={font=\footnotesize},
        tick label style={font=\footnotesize},
    ]
        
    \addplot[line width=0.25mm,mark size=2pt,mark=o, color=B6] 
        plot coordinates {
(0, 0.926345706)
(1, 0.924990177)
(2, 0.924781442)
(3, 0.926929235)
(4, 0.925914884)
(5, 0.924817562)
(6, 0.924767494)
(7, 0.923572898)
(8, 0.921267688)
(9, 0.919164538)
(10, 0.911055565)
};
 \addplot[line width=0.25mm,mark size=2pt,mark=square, color=B2] 
        plot coordinates {
(0, 0.925895452)
(1, 0.923720837)
(2, 0.925268352)
(3, 0.925468206)
(4, 0.925997376)
(5, 0.924856901)
(6, 0.922854662)
(7, 0.923377991)
(8, 0.922290087)
(9, 0.918675184)
(10, 0.910365641)
};
    \addplot[line width=0.25mm,mark size=2pt,mark=pentagon, color=O3] 
        plot coordinates {
(0, 0.926538527)
(1, 0.925812066)
(2, 0.924463749)
(3, 0.924845457)
(4, 0.925284624)
(5, 0.925196111)
(6, 0.923774362)
(7, 0.92316705)
(8, 0.921840489)
(9, 0.920623779)
(10, 0.909796476)
};
    \end{axis}
\end{tikzpicture}\hspace{0mm}%
}%
\subfloat[{MLM}]{
\begin{tikzpicture}[scale=1]\label{subfig:beta_MLM}
    \begin{axis}[
        height=\columnwidth/2.7,
        width=\columnwidth/2.3,
        ylabel={\em AUC},
        xmin=0, xmax=10,
        ymin=0.802, ymax=0.810,
        xtick={1,3,5,7,9},
        xticklabel style = {font=\footnotesize},
        xticklabels={0.1,0.3,0.5,0.7,0.9},
        scaled y ticks = false,
        yticklabel style={/pgf/number format/fixed zerofill,/pgf/number format/precision=3},
        every axis y label/.style={at={(current axis.north west)},right=3mm,above=0mm},
        label style={font=\footnotesize},
        tick label style={font=\footnotesize},
    ]
        
    \addplot[line width=0.25mm,mark size=2pt,mark=o, color=B6] 
        plot coordinates {
(0, 0.806901515)
(1, 0.808486819)
(2, 0.807086229)
(3, 0.807953477)
(4, 0.808726251)
(5, 0.807393849)
(6, 0.806926608)
(7, 0.804309487)
(8, 0.804242373)
(9, 0.798984766)
(10, 0.7751652)
};
 \addplot[line width=0.25mm,mark size=2pt,mark=square, color=B2] 
        plot coordinates {
(0, 0.806059599)
(1, 0.80937922)
(2, 0.806318164)
(3, 0.806173086)
(4, 0.806500554)
(5, 0.803555071)
(6, 0.803437352)
(7, 0.80187887)
(8, 0.800995111)
(9, 0.795158684)
(10, 0.77147454)
};
    \addplot[line width=0.25mm,mark size=2pt,mark=pentagon, color=O3] 
        plot coordinates {
(0, 0.80658567)
(1, 0.807298481)
(2, 0.806726992)
(3, 0.805548072)
(4, 0.805461466)
(5, 0.805856586)
(6, 0.803315401)
(7, 0.803436518)
(8, 0.802298784)
(9, 0.79387331)
(10, 0.773126364)
};
    \end{axis}
\end{tikzpicture}\hspace{0mm}%
}%
\subfloat[{Review}]{
\begin{tikzpicture}[scale=1]\label{subfig:beta_Review}
    \begin{axis}[
        height=\columnwidth/2.7,
        width=\columnwidth/2.3,
        ylabel={\em AUC},
        xmin=0, xmax=10,
        ymin=0.710, ymax=0.775,
        xtick={1,3,5,7,9},
        xticklabel style = {font=\footnotesize},
        xticklabels={0.1,0.3,0.5,0.7,0.9},
        scaled y ticks = false,
        yticklabel style={/pgf/number format/fixed zerofill,/pgf/number format/precision=2},
        every axis y label/.style={at={(current axis.north west)},right=3mm,above=0mm},
        label style={font=\footnotesize},
        tick label style={font=\footnotesize},
    ]
        
    \addplot[line width=0.25mm,mark size=2pt,mark=o, color=B6] 
        plot coordinates {
(0, 0.702694416)
(1, 0.708205819)
(2, 0.774035573)
(3, 0.717697501)
(4, 0.738518119)
(5, 0.716472745)
(6, 0.682486236)
(7, 0.676668704)
(8, 0.654929638)
(9, 0.693815112)
(10, 0.699020207)
};
 \addplot[line width=0.25mm,mark size=2pt,mark=square, color=B2] 
        plot coordinates {
(0, 0.751377821)
(1, 0.733466029)
(2, 0.72473979)
(3, 0.763625264)
(4, 0.744335592)
(5, 0.705143929)
(6, 0.680036783)
(7, 0.682486176)
(8, 0.629210114)
(9, 0.683864117)
(10, 0.691977978)
};
    \addplot[line width=0.25mm,mark size=2pt,mark=pentagon, color=O3] 
        plot coordinates {
(0, 0.720759332)
(1, 0.745560348)
(2, 0.714329481)
(3, 0.713717103)
(4, 0.740661383)
(5, 0.714635611)
(6, 0.671769738)
(7, 0.683098555)
(8, 0.644519269)
(9, 0.709124327)
(10, 0.704225361)
};
    \end{axis}
\end{tikzpicture}\hspace{0mm}%
}%
\subfloat[{Senate}]{
\begin{tikzpicture}[scale=1]\label{subfig:beta_Senate}
    \begin{axis}[
        height=\columnwidth/2.7,
        width=\columnwidth/2.3,
        ylabel={\em AUC},
        xmin=0, xmax=10,
        ymin=0.897, ymax=0.905,
        xtick={1,3,5,7,9},
        xticklabel style = {font=\footnotesize},
        xticklabels={0.1,0.3,0.5,0.7,0.9},
        scaled y ticks = false,
        yticklabel style={/pgf/number format/fixed zerofill,/pgf/number format/precision=3},
        every axis y label/.style={at={(current axis.north west)},right=3mm,above=0mm},
        label style={font=\footnotesize},
        tick label style={font=\footnotesize},
    ]
        
    \addplot[line width=0.25mm,mark size=2pt,mark=o, color=B6] 
        plot coordinates {
(0, 0.89721179)
(1, 0.90035212)
(2, 0.900656819)
(3, 0.901304841)
(4, 0.899005473)
(5, 0.901125908)
(6, 0.898722768)
(7, 0.894321442)
(8, 0.897195876)
(9, 0.898292422)
(10, 0.900099218)
};
 \addplot[line width=0.25mm,mark size=2pt,mark=square, color=B2] 
        plot coordinates {
(0, 0.89899826)
(1, 0.897935867)
(2, 0.900719345)
(3, 0.904486597)
(4, 0.902830541)
(5, 0.901794374)
(6, 0.900475144)
(7, 0.894004822)
(8, 0.901737869)
(9, 0.903193593)
(10, 0.896463692)
};
    \addplot[line width=0.25mm,mark size=2pt,mark=pentagon, color=O3] 
        plot coordinates {
(0, 0.90316987)
(1, 0.901876509)
(2, 0.900216579)
(3, 0.9000808)
(4, 0.897793353)
(5, 0.901052475)
(6, 0.900719106)
(7, 0.897358537)
(8, 0.901507676)
(9, 0.903716981)
(10, 0.892710745)
};
    \end{axis}
\end{tikzpicture}\hspace{0mm}%
}%
\end{small}
\vspace{-3mm}
\caption{AUC by varying $\mu$ in \algo.} \label{fig:beta}
\vspace{-2mm}
\end{figure*}

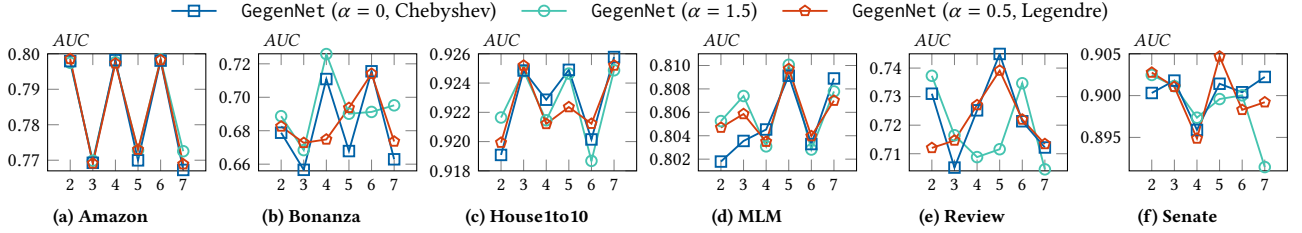
\begin{figure*}[!t]
\begin{small}
\begin{tikzpicture}
    \begin{customlegend}[legend columns=3,
        legend entries={{\algo{} (\(\alpha=0\), Chebyshev)},{\algo{} (\(\alpha=1.5\))}, {\algo{} (\(\alpha=0.5\), Legendre)}},
        legend style={at={(0.45,1.15)},anchor=north,draw=none,font=\small,column sep=0.25cm}]
    \addlegendimage{line width=0.25mm,mark size=2pt,mark=square, color=B2}
    \addlegendimage{line width=0.25mm,mark size=2pt,mark=o, color=B6}
    \addlegendimage{line width=0.25mm,mark size=2pt,mark=pentagon, color=O3}
    \end{customlegend}
\end{tikzpicture}
\\[-\lineskip]
\vspace{-4mm}
\hspace*{0mm}
\subfloat[{Amazon}]{
\begin{tikzpicture}[scale=1]\label{subfig:layer_Amazon}
    \begin{axis}[
        height=\columnwidth/2.7,
        width=\columnwidth/2.3,
        ylabel={\em AUC},
        xmin=-1, xmax=6,
        ymin=0.767, ymax=0.8,
        xtick={0,1,2,3,4,5},
        xticklabel style = {font=\footnotesize},
        xticklabels={2,3,4,5,6,7},
        scaled y ticks = false,
        yticklabel style={/pgf/number format/fixed zerofill,/pgf/number format/precision=2},
        every axis y label/.style={at={(current axis.north west)},right=3mm,above=0mm},
        label style={font=\footnotesize},
        tick label style={font=\footnotesize},
    ]
        
    \addplot[line width=0.25mm,mark size=2pt,mark=o, color=B6] 
        plot coordinates {
(0, 0.797469735)
(1, 0.769871056)
(2, 0.797621489)
(3, 0.772793233)
(4, 0.798489809)
(5, 0.772535384)
};
 \addplot[line width=0.25mm,mark size=2pt,mark=square, color=B2] 
        plot coordinates {
(0, 0.797932148)
(1, 0.76927042)
(2, 0.798211157)
(3, 0.769963324)
(4, 0.798083186)
(5, 0.767261982)
};
    \addplot[line width=0.25mm,mark size=2pt,mark=pentagon, color=O3] 
        plot coordinates {
(0, 0.798558891)
(1, 0.769093633)
(2, 0.797294021)
(3, 0.772981167)
(4, 0.798247218)
(5, 0.76886785)
};
    \end{axis}
\end{tikzpicture}\hspace{0mm}%
}%
\subfloat[{Bonanza}]{
\begin{tikzpicture}[scale=1]\label{subfig:layer_Bonanza}
    \begin{axis}[
        height=\columnwidth/2.7,
        width=\columnwidth/2.3,
        ylabel={\em AUC},
        xmin=-1, xmax=6,
        ymin=0.656, ymax=0.726,
        xtick={0,1,2,3,4,5},
        xticklabel style = {font=\footnotesize},
        xticklabels={2,3,4,5,6,7},
        scaled y ticks = false,
        yticklabel style={/pgf/number format/fixed zerofill,/pgf/number format/precision=2},
        every axis y label/.style={at={(current axis.north west)},right=3mm,above=0mm},
        label style={font=\footnotesize},
        tick label style={font=\footnotesize},
    ]
        
    \addplot[line width=0.25mm,mark size=2pt,mark=o, color=B6] 
        plot coordinates {
(0, 0.688699245)
(1, 0.668401837)
(2, 0.725830674)
(3, 0.690153062)
(4, 0.691232145)
(5, 0.695181191)
};
 \addplot[line width=0.25mm,mark size=2pt,mark=square, color=B2] 
        plot coordinates {
(0, 0.678809643)
(1, 0.656624675)
(2, 0.710900247)
(3, 0.667837262)
(4, 0.715534151)
(5, 0.662909448)
};
    \addplot[line width=0.25mm,mark size=2pt,mark=pentagon, color=O3] 
        plot coordinates {
(0, 0.682556093)
(1, 0.672748804)
(2, 0.674788892)
(3, 0.693675518)
(4, 0.714085698)
(5, 0.673560858)
};
    \end{axis}
\end{tikzpicture}\hspace{0mm}%
}%
\subfloat[{House1to10}]{
\begin{tikzpicture}[scale=1]\label{subfig:layer_House1to10}
    \begin{axis}[
        height=\columnwidth/2.7,
        width=\columnwidth/2.3,
        ylabel={\em AUC},
        xmin=-1, xmax=6,
        ymin=0.918, ymax=0.926,
        xtick={0,1,2,3,4,5},
        xticklabel style = {font=\footnotesize},
        xticklabels={2,3,4,5,6,7},
        scaled y ticks = false,
        yticklabel style={/pgf/number format/fixed zerofill,/pgf/number format/precision=3},
        every axis y label/.style={at={(current axis.north west)},right=3mm,above=0mm},
        label style={font=\footnotesize},
        tick label style={font=\footnotesize},
    ]
        
    \addplot[line width=0.25mm,mark size=2pt,mark=o, color=B6] 
        plot coordinates {
(0, 0.92164582)
(1, 0.924817562)
(2, 0.921456814)
(3, 0.924647808)
(4, 0.918682337)
(5, 0.924873471)
};
 \addplot[line width=0.25mm,mark size=2pt,mark=square, color=B2] 
        plot coordinates {
(0, 0.91908884)
(1, 0.924856901)
(2, 0.922861814)
(3, 0.924904525)
(4, 0.920147061)
(5, 0.925779104)
};
    \addplot[line width=0.25mm,mark size=2pt,mark=pentagon, color=O3] 
        plot coordinates {
(0, 0.919929326)
(1, 0.925196111)
(2, 0.921206355)
(3, 0.922359467)
(4, 0.921222806)
(5, 0.925200522)
};
    \end{axis}
\end{tikzpicture}\hspace{0mm}%
}%
\subfloat[{MLM}]{
\begin{tikzpicture}[scale=1]\label{subfig:layer_MLM}
    \begin{axis}[
        height=\columnwidth/2.7,
        width=\columnwidth/2.3,
        ylabel={\em AUC},
        xmin=-1, xmax=6,
        ymin=0.801, ymax=0.811,
        xtick={0,1,2,3,4,5},
        xticklabel style = {font=\footnotesize},
        xticklabels={2,3,4,5,6,7},
        scaled y ticks = false,
        yticklabel style={/pgf/number format/fixed zerofill,/pgf/number format/precision=3},
        every axis y label/.style={at={(current axis.north west)},right=3mm,above=0mm},
        label style={font=\footnotesize},
        tick label style={font=\footnotesize},
    ]
        
    \addplot[line width=0.25mm,mark size=2pt,mark=o, color=B6] 
        plot coordinates {
(0, 0.805253029)
(1, 0.807393849)
(2, 0.803091586)
(3, 0.810061097)
(4, 0.802821875)
(5, 0.807783484)
};
 \addplot[line width=0.25mm,mark size=2pt,mark=square, color=B2] 
        plot coordinates {
(0, 0.801794171)
(1, 0.803555071)
(2, 0.804532647)
(3, 0.809126258)
(4, 0.803289771)
(5, 0.808892548)
};
    \addplot[line width=0.25mm,mark size=2pt,mark=pentagon, color=O3] 
        plot coordinates {
(0, 0.804694533)
(1, 0.805856586)
(2, 0.803490996)
(3, 0.809704542)
(4, 0.803970039)
(5, 0.80699718)
};
    \end{axis}
\end{tikzpicture}\hspace{0mm}%
}%
\subfloat[{Review}]{
\begin{tikzpicture}[scale=1]\label{subfig:layer_Review}
    \begin{axis}[
        height=\columnwidth/2.7,
        width=\columnwidth/2.3,
        ylabel={\em AUC},
        xmin=-1, xmax=6,
        ymin=0.704, ymax=0.745,
        xtick={0,1,2,3,4,5},
        xticklabel style = {font=\footnotesize},
        xticklabels={2,3,4,5,6,7},
        scaled y ticks = false,
        yticklabel style={/pgf/number format/fixed zerofill,/pgf/number format/precision=2},
        every axis y label/.style={at={(current axis.north west)},right=3mm,above=0mm},
        label style={font=\footnotesize},
        tick label style={font=\footnotesize},
    ]
        
    \addplot[line width=0.25mm,mark size=2pt,mark=o, color=B6] 
        plot coordinates {
(0, 0.737293363)
(1, 0.716472745)
(2, 0.708818138)
(3, 0.71157378)
(4, 0.734690785)
(5, 0.70453155)
};
 \addplot[line width=0.25mm,mark size=2pt,mark=square, color=B2] 
        plot coordinates {
(0, 0.731016576)
(1, 0.705143929)
(2, 0.725199044)
(3, 0.74494797)
(4, 0.72137177)
(5, 0.712186158)
};
    \addplot[line width=0.25mm,mark size=2pt,mark=pentagon, color=O3] 
        plot coordinates {
(0, 0.712033093)
(1, 0.714635611)
(2, 0.727036119)
(3, 0.739130437)
(4, 0.721984088)
(5, 0.713410854)
};
    \end{axis}
\end{tikzpicture}\hspace{0mm}%
}%
\subfloat[{Senate}]{
\begin{tikzpicture}[scale=1]\label{subfig:layer_Senate}
    \begin{axis}[
        height=\columnwidth/2.7,
        width=\columnwidth/2.3,
        ylabel={\em AUC},
        xmin=-1, xmax=6,
        ymin=0.891, ymax=0.905,
        xtick={0,1,2,3,4,5},
        xticklabel style = {font=\footnotesize},
        xticklabels={2,3,4,5,6,7},
        scaled y ticks = false,
        yticklabel style={/pgf/number format/fixed zerofill,/pgf/number format/precision=3},
        every axis y label/.style={at={(current axis.north west)},right=3mm,above=0mm},
        label style={font=\footnotesize},
        tick label style={font=\footnotesize},
    ]
        
    \addplot[line width=0.25mm,mark size=2pt,mark=o, color=B6] 
        plot coordinates {
(0, 0.902467787)
(1, 0.901125908)
(2, 0.89735961)
(3, 0.899586737)
(4, 0.900032163)
(5, 0.891453147)
};
 \addplot[line width=0.25mm,mark size=2pt,mark=square, color=B2] 
        plot coordinates {
(0, 0.900313318)
(1, 0.901794374)
(2, 0.895895243)
(3, 0.901417434)
(4, 0.900403023)
(5, 0.902233124)
};
    \addplot[line width=0.25mm,mark size=2pt,mark=pentagon, color=O3] 
        plot coordinates {
(0, 0.902758479)
(1, 0.901052475)
(2, 0.894870281)
(3, 0.904678822)
(4, 0.898297966)
(5, 0.899206758)
};
    \end{axis}
\end{tikzpicture}\hspace{0mm}%
}%
\end{small}
\vspace{-3mm}
\caption{AUC by varying $L$ in \algo.} \label{fig:layer}
\vspace{-2mm}
\end{figure*}

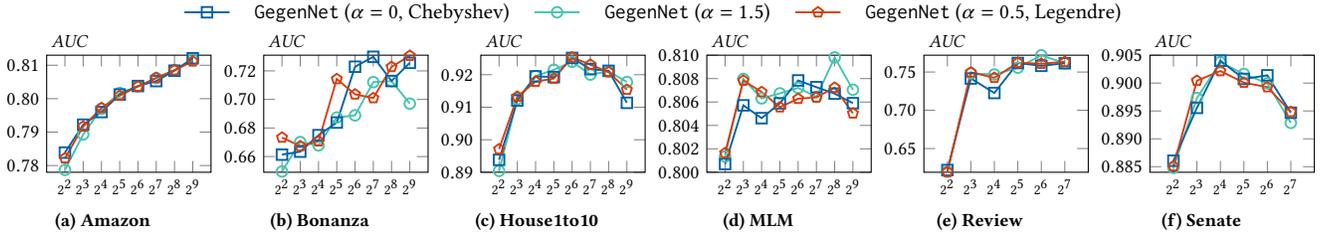
\begin{figure*}[!t]
\begin{small}
\begin{tikzpicture}
    \begin{customlegend}[legend columns=3,
        legend entries={{\algo{} (\(\alpha=0\), Chebyshev)},{\algo{} (\(\alpha=1.5\))}, {\algo{} (\(\alpha=0.5\), Legendre)}},
        legend style={at={(0.45,1.15)},anchor=north,draw=none,font=\small,column sep=0.25cm}]
    \addlegendimage{line width=0.25mm,mark size=2pt,mark=square, color=B2}
    \addlegendimage{line width=0.25mm,mark size=2pt,mark=o, color=B6}
    \addlegendimage{line width=0.25mm,mark size=2pt,mark=pentagon, color=O3}
    \end{customlegend}
\end{tikzpicture}
\\[-\lineskip]
\vspace{-4mm}
\hspace*{0mm}
\subfloat[{Amazon}]{
\begin{tikzpicture}[scale=1]\label{subfig:svd_Amazon}
    \begin{axis}[
        height=\columnwidth/2.7,
        width=\columnwidth/2.25,
        ylabel={\em AUC},
        xmin=-1, xmax=8,
        ymin=0.778, ymax=0.813,
        xtick={0,1,2,3,4,5,6,7},
        xticklabel style = {font=\tiny},
        xticklabels={$2^2$,$2^3$,$2^4$,$2^5$,$2^6$,$2^7$,$2^8$,$2^9$},
        scaled y ticks = false,
        yticklabel style={/pgf/number format/fixed zerofill,/pgf/number format/precision=2},
        every axis y label/.style={at={(current axis.north west)},right=3mm,above=0mm},
        label style={font=\footnotesize},
        tick label style={font=\footnotesize},
    ]
        
    \addplot[line width=0.25mm,mark size=2pt,mark=o, color=B6] 
        plot coordinates {
(0, 0.778707266)
(1, 0.789313793)
(2, 0.796953201)
(3, 0.80169934)
(4, 0.803700328)
(5, 0.806176066)
(6, 0.808554173)
(7, 0.811681747)
};
 \addplot[line width=0.25mm,mark size=2pt,mark=square, color=B2] 
        plot coordinates {
(0, 0.783870101)
(1, 0.792217612)
(2, 0.796007574)
(3, 0.801243901)
(4, 0.803779125)
(5, 0.805264115)
(6, 0.808366776)
(7, 0.812083364)
};
    \addplot[line width=0.25mm,mark size=2pt,mark=pentagon, color=O3] 
        plot coordinates {
(0, 0.782192588)
(1, 0.791683495)
(2, 0.797212362)
(3, 0.800898552)
(4, 0.803653717)
(5, 0.806198061)
(6, 0.808515728)
(7, 0.811150074)
};
    \end{axis}
\end{tikzpicture}\hspace{0mm}%
}%
\subfloat[{Bonanza}]{
\begin{tikzpicture}[scale=1]\label{subfig:svd_Bonanza}
    \begin{axis}[
        height=\columnwidth/2.7,
        width=\columnwidth/2.25,
        ylabel={\em AUC},
        xmin=-1, xmax=8,
        ymin=0.649, ymax=0.731,
        xtick={0,1,2,3,4,5,6,7},
        xticklabel style = {font=\tiny},
        xticklabels={$2^2$,$2^3$,$2^4$,$2^5$,$2^6$,$2^7$,$2^8$,$2^9$},
        scaled y ticks = false,
        yticklabel style={/pgf/number format/fixed zerofill,/pgf/number format/precision=2},
        every axis y label/.style={at={(current axis.north west)},right=3mm,above=0mm},
        label style={font=\footnotesize},
        tick label style={font=\footnotesize},
    ]
        
    \addplot[line width=0.25mm,mark size=2pt,mark=o, color=B6] 
        plot coordinates {
(0, 0.649565458)
(1, 0.670142531)
(2, 0.667822897)
(3, 0.687327921)
(4, 0.688848019)
(5, 0.712131798)
(6, 0.71281296)
(7, 0.697072387)
};
 \addplot[line width=0.25mm,mark size=2pt,mark=square, color=B2] 
        plot coordinates {
(0, 0.661335468)
(1, 0.66342926)
(2, 0.674978852)
(3, 0.683812618)
(4, 0.722915947)
(5, 0.729862213)
(6, 0.712913394)
(7, 0.725651443)
};
    \addplot[line width=0.25mm,mark size=2pt,mark=pentagon, color=O3] 
        plot coordinates {
(0, 0.673327923)
(1, 0.667310178)
(2, 0.67067486)
(3, 0.714288235)
(4, 0.703572273)
(5, 0.70094794)
(6, 0.722783267)
(7, 0.730629444)
};
    \end{axis}
\end{tikzpicture}\hspace{0mm}%
}%
\subfloat[{House1to10}]{
\begin{tikzpicture}[scale=1]\label{subfig:svd_House1to10}
    \begin{axis}[
        height=\columnwidth/2.7,
        width=\columnwidth/2.25,
        ylabel={\em AUC},
        xmin=-1, xmax=8,
        ymin=0.890, ymax=0.926,
        xtick={0,1,2,3,4,5,6,7},
        xticklabel style = {font=\tiny},
        xticklabels={$2^2$,$2^3$,$2^4$,$2^5$,$2^6$,$2^7$,$2^8$,$2^9$},
        scaled y ticks = false,
        yticklabel style={/pgf/number format/fixed zerofill,/pgf/number format/precision=2},
        every axis y label/.style={at={(current axis.north west)},right=3mm,above=0mm},
        label style={font=\footnotesize},
        tick label style={font=\footnotesize},
    ]
        
    \addplot[line width=0.25mm,mark size=2pt,mark=o, color=B6] 
        plot coordinates {
(0, 0.890417337)
(1, 0.912513971)
(2, 0.919580817)
(3, 0.921505094)
(4, 0.923971534)
(5, 0.920032799)
(6, 0.921025395)
(7, 0.917754352)
};
 \addplot[line width=0.25mm,mark size=2pt,mark=square, color=B2] 
        plot coordinates {
(0, 0.89381671)
(1, 0.912068605)
(2, 0.919492722)
(3, 0.919250369)
(4, 0.925159335)
(5, 0.921716452)
(6, 0.921195209)
(7, 0.911344647)
};
    \addplot[line width=0.25mm,mark size=2pt,mark=pentagon, color=O3] 
        plot coordinates {
(0, 0.89709568)
(1, 0.913254797)
(2, 0.917869329)
(3, 0.918807566)
(4, 0.925366282)
(5, 0.923101604)
(6, 0.920799971)
(7, 0.91539669)
};
    \end{axis}
\end{tikzpicture}\hspace{0mm}%
}%
\subfloat[{MLM}]{
\begin{tikzpicture}[scale=1]\label{subfig:svd_MLM}
    \begin{axis}[
        height=\columnwidth/2.7,
        width=\columnwidth/2.25,
        ylabel={\em AUC},
        xmin=-1, xmax=8,
        ymin=0.800, ymax=0.810,
        xtick={0,1,2,3,4,5,6,7},
        xticklabel style = {font=\tiny},
        xticklabels={$2^2$,$2^3$,$2^4$,$2^5$,$2^6$,$2^7$,$2^8$,$2^9$},
        scaled y ticks = false,
        yticklabel style={/pgf/number format/fixed zerofill,/pgf/number format/precision=3},
        every axis y label/.style={at={(current axis.north west)},right=3mm,above=0mm},
        label style={font=\footnotesize},
        tick label style={font=\footnotesize},
    ]
        
    \addplot[line width=0.25mm,mark size=2pt,mark=o, color=B6] 
        plot coordinates {
(0, 0.801343262)
(1, 0.807974815)
(2, 0.806311488)
(3, 0.806738138)
(4, 0.807204783)
(5, 0.806435347)
(6, 0.809810162)
(7, 0.80704695)
};
 \addplot[line width=0.25mm,mark size=2pt,mark=square, color=B2] 
        plot coordinates {
(0, 0.800700426)
(1, 0.805715799)
(2, 0.804619193)
(3, 0.80588907)
(4, 0.807842195)
(5, 0.807268858)
(6, 0.806715488)
(7, 0.805898786)
};
    \addplot[line width=0.25mm,mark size=2pt,mark=pentagon, color=O3] 
        plot coordinates {
(0, 0.801638842)
(1, 0.807834387)
(2, 0.806856155)
(3, 0.80555582)
(4, 0.806278408)
(5, 0.806396663)
(6, 0.807224751)
(7, 0.805031419)
};
    \end{axis}
\end{tikzpicture}\hspace{0mm}%
}%
\subfloat[{Review}]{
\begin{tikzpicture}[scale=1]\label{subfig:svd_Review}
    \begin{axis}[
        height=\columnwidth/2.7,
        width=\columnwidth/2.25,
        ylabel={\em AUC},
        xmin=-1, xmax=6,
        ymin=0.619, ymax=0.772,
        xtick={0,1,2,3,4,5},
        xticklabel style = {font=\tiny},
        xticklabels={$2^2$,$2^3$,$2^4$,$2^5$,$2^6$,$2^7$},
        scaled y ticks = false,
        yticklabel style={/pgf/number format/fixed zerofill,/pgf/number format/precision=2},
        every axis y label/.style={at={(current axis.north west)},right=3mm,above=0mm},
        label style={font=\footnotesize},
        tick label style={font=\footnotesize},
    ]
        
    \addplot[line width=0.25mm,mark size=2pt,mark=o, color=B6] 
        plot coordinates {
(0, 0.619718313)
(1, 0.74831599)
(2, 0.746785045)
(3, 0.755664468)
(4, 0.77189225)
(5, 0.761481941)
};
 \addplot[line width=0.25mm,mark size=2pt,mark=square, color=B2] 
        plot coordinates {
(0, 0.621861577)
(1, 0.74157995)
(2, 0.722596467)
(3, 0.76178807)
(4, 0.758113861)
(5, 0.761482)
};
    \addplot[line width=0.25mm,mark size=2pt,mark=pentagon, color=O3] 
        plot coordinates {
(0, 0.619412124)
(1, 0.749540746)
(2, 0.742498517)
(3, 0.762094259)
(4, 0.760869622)
(5, 0.763319075)
};
    \end{axis}
\end{tikzpicture}\hspace{0mm}%
}%
\subfloat[{Senate}]{
\begin{tikzpicture}[scale=1]\label{subfig:svd_Senate}
    \begin{axis}[
        height=\columnwidth/2.7,
        width=\columnwidth/2.25,
        ylabel={\em AUC},
        xmin=-1, xmax=6,
        ymin=0.884, ymax=0.905,
        xtick={0,1,2,3,4,5},
        xticklabel style = {font=\tiny},
        xticklabels={$2^2$,$2^3$,$2^4$,$2^5$,$2^6$,$2^7$},
        scaled y ticks = false,
        yticklabel style={/pgf/number format/fixed zerofill,/pgf/number format/precision=3},
        every axis y label/.style={at={(current axis.north west)},right=3mm,above=0mm},
        label style={font=\footnotesize},
        tick label style={font=\footnotesize},
    ]
        
    \addplot[line width=0.25mm,mark size=2pt,mark=o, color=B6] 
        plot coordinates {
(0, 0.884781778)
(1, 0.897322118)
(2, 0.903716385)
(3, 0.901633978)
(4, 0.899956286)
(5, 0.892854333)
};
 \addplot[line width=0.25mm,mark size=2pt,mark=square, color=B2] 
        plot coordinates {
(0, 0.886069357)
(1, 0.895554423)
(2, 0.904011846)
(3, 0.900741935)
(4, 0.901385844)
(5, 0.894681334)
};
    \addplot[line width=0.25mm,mark size=2pt,mark=pentagon, color=O3] 
        plot coordinates {
(0, 0.885089159)
(1, 0.900421739)
(2, 0.902160466)
(3, 0.900072455)
(4, 0.89926815)
(5, 0.894809365)
};
    \end{axis}
\end{tikzpicture}\hspace{0mm}%
}%
\end{small}
\vspace{-3mm}
\caption{AUC by varying $d$ in \algo.} \label{fig:svd}
\vspace{-2mm}
\end{figure*}

This section empirically evaluates the effectiveness of our proposed algorithm in link sign classification on six real SBG datasets~\cite{huang2021signed,zhang2023contrastive} from various application domains.
\eat{
The Amazon dataset contains book ratings, with edges representing user reviews labeled as positive or negative. 
Bonanza, from an e-commerce platform, includes user ratings of sellers as "Positive," "Neutral," or "Negative," with neutral scores treated as ambiguous. 
House1to10 and Senate record voting behaviors in the U.S. House of Representatives and Senate, respectively, where edges correspond to “Yea” (positive) or “Nay” (negative) votes on congressional bills. 
The MLM (MovieLens Movies) dataset captures user ratings of movies, while the Review dataset consists of peer review decisions from a top computer science conference, with edges labeled as "Accept" (positive) or "Reject" (negative).
}
The statistics of the datasets are summarized in Table \ref{tbl:datasets}. 
We randomly split all edges into training, cross-validation, and test sets with an $8:1:1$ ratio.
For a robust and fair evaluation, all experiments are conducted on a Linux machine equipped with 4 AMD EPYC 7313 CPUs (500GB RAM) and 1 NVIDIA RTX A5000 GPU (24GB memory). The codebase and datasets are publicly available at \url{https://github.com/wanghewen/GegenNet}.

\subsection{Baselines and Hyperparameters}

We compare \algo{} against 11 competitors, including \textsf{MSGNN} \cite{he2022msgnn}, \textsf{SDGNN} \cite{huang2021sdgnn}, \textsf{SGCN} \cite{derr2018signed}, \textsf{SiGAT} \cite{huang2019signed}, \textsf{SigMaNet} \cite{10.1609/aaai.v37i6.25919}, \textsf{SNEA} \cite{li2020learning}, \textsf{SSSNET} \cite{he2022sssnet}, \textsf{SBGCL} \cite{zhang2023contrastive}, \textsf{SBGNN} \cite{huang2021signed}, \textsf{SidNet} \cite{jung2022signed}, \textsf{SLGNN} \cite{li2023signed}, in terms of edge classification accuracy. For \textsf{MSGNN}, \textsf{SDGNN}, \textsf{SGCN}, \textsf{SiGAT}, \textsf{SigMaNet}, \textsf{SNEA} and \textsf{SSSNET}, we leverage the implementation in~\cite{he2022msgnn}.
For \textsf{SBGCL}, \textsf{SBGNN}, \textsf{SidNet}, and \textsf{SLGNN}, we use the source codes from
the respective authors. The embedding size of the node is set to 32, and the maximum training epochs to 300 in different methods. For the rest of the parameters, we adopt the parameter settings suggested in their respective papers. For \algo, we set the learning rate to be 0.01, dropout rate to be 0.5, weight decay to be 1e-5, weight parameter $\mu$ to be 0.3, the number of model layers $L$ to be 3, and SVD dimension $d$ to be 32.

\subsection{Link Sign Prediction Performance}
\looseness=-1 We evaluate the performance of \algo on the link sign prediction task, with results presented in Table \ref{tbl:classification-perf} in terms of AUC and F1-score. \algo consistently outperforms existing methods across all datasets, effectively capturing both the structural properties and sign dynamics inherent to signed bipartite graphs. For datasets where certain methods encounter errors during execution or fail to converge, the results are excluded and marked as “-” in Table \ref{tbl:classification-perf}.

\algo achieves significant improvements over baseline models. On all six datasets, \algo yields the highest AUC and F1 scores. Such results highlight the model’s robustness in learning nuanced positive and negative relationships while preserving the bipartite structure. A key factor contributing to \algo’s success is its novel spectral filtering mechanism based on Gegenbauer polynomials, which offers better expressiveness for both low- and high-frequency graph signals. Compared to alternative spectral bases like Chebyshev or Legendre polynomials, the Gegenbauer-based filtering with $\alpha=1.5$ consistently demonstrates superior adaptability to the graph signal distributions.

Furthermore, \algo shows strong resilience on sparse and imbalanced datasets. For example, the Bonanza dataset contains very few negative links, while the Review dataset is characterized by sparsely connected nodes. Despite these challenges, \algo maintains robust performance, achieving competitive AUC and F1 scores where other methods tend to struggle. 
Overall, \algo's design effectively preserves the bipartite graph structure and distinguishes between positive and negative relationships during spectral convolutions. By modeling both inter- and intra-partition proximities, \algo captures comprehensive graph information, leading to significant improvements in link sign prediction. 
\raggedbottom
\subsection{Ablation Study}
Table \ref{tbl:ablation} presents the ablation study results, evaluating the impact of spectral transformations (ST), as well as random embeddings, on the performance of the \algo model. 
The full \algo model, incorporating both positive and negative spectral transformations, consistently achieves superior performance. For instance, in the Amazon dataset, the Chebyshev-based model attains an AUC of 0.8006 and an F1 score of 0.6834. Omitting the negative ST results in a significant decline to an AUC of 0.7687 and an F1 score of 0.6643. Similarly, in the Bonanza dataset, the AUC of model with $\alpha=1.5$ decreases from 0.7301 to 0.6500 upon removal of the negative ST. Notably, the negative spectral transformation has a more pronounced effect on performance. In the Bonanza dataset, with $\alpha=1.5$, eliminating the negative ST leads to a sharp AUC reduction from 0.7301 to 0.6500, underscoring its critical role in capturing structural and signed relationships within the graph. In contrast, removing the positive ST results in a smaller performance decline. Introducing random embeddings, devoid of spectral transformations, leads to the most severe performance degradation across all datasets. In the MLM dataset, substituting learned embeddings with random ones reduces the AUC from 0.8074 to 0.7324.
In summary, both positive and negative spectral transformations are vital for optimal performance. Replacing spectral transformations with random embeddings substantially degrades performance, emphasizing the importance of spectral learning in capturing meaningful graph structures and enhancing link sign prediction accuracy.
\raggedbottom

\subsection{Hyperparameter Analysis}
Figure \ref{fig:beta} demonstrates that the choice of $\mu$ and the type of basis function significantly impact the model's performance (as measured by AUC). There is a consistent trend of performance degradation as $\mu$ increases beyond 0.7 for all datasets and basis functions. The results suggest that optimal $\mu$ values are dataset-dependent, but generally lie between 0.1 and 0.5 across different datasets.

Figure \ref{fig:layer} shows that different datasets exhibit varying levels of sensitivity to changes in the number of transformation layers in \algo. For Amazon, Bonanza, and Review, the AUC shows significant fluctuations, while House1to10, MLM, and Senate exhibit more stable behavior. When $\alpha=1.5$, model tends to outperform the other two basis functions in datasets such as Bonanza and MLM, but Chebyshev is able to achieve the best results in datasets like House1to10 and Review. The optimal number of transformation layers varies by dataset and basis function, underscoring the importance of dataset-specific tuning.

Figure \ref{fig:svd} illustrates how model performance changes as the SVD dimension increases. Note that the number of nodes must be less than the SVD dimension, so in the Review and Senate datasets, the SVD dimension is truncated to 128. Across all datasets, increasing the SVD dimension generally leads to improved AUC, with the highest performance typically observed between \(2^4\) and \(2^7\). In Amazon, Bonanza, and Review, performance consistently improves with higher SVD dimensions, while in House1to10, MLM, and Senate, performance stabilizes after a certain point, with only minor gains beyond that. These results emphasize the importance of selecting an appropriate SVD dimension for each dataset, as overfitting may occur at higher dimensions in some cases.

\section{Conclusion}
This paper presents \algo, a spectral convolutional neural network tailored for signed bipartite graphs. By leveraging Gegenbauer polynomials for spectral filtering, \algo effectively captures both low- and high-frequency graph signals. Experiments across six real-world datasets demonstrate its superiority in link sign prediction tasks compared to existing models. Ablation studies further confirm the critical role of both positive and negative spectral transformations in enhancing performance. Future work may explore extending \algo to dynamic or multiplex bipartite networks to handle temporal information and multiple interaction types.

\begin{acks}
This research is supported by the Ministry of Education, Singapore, under its MOE AcRF TIER 3 Grant (MOE-MOET32022-0001). Renchi Yang is supported by the National Natural Science Foundation of China (No. 62302414), the Hong Kong RGC ECS grant (No. 22202623), and the Huawei Gift Fund.
\end{acks}

\appendix

\section{Theoretical Proofs}\label{sec:proof}

\begin{proof}[\bf Proof of Proposition~\ref{prop:linpred}]
First, according to the definition of $\UM$ and $\boldsymbol{\Lambda}$, we can derive
\(\AM^2 = \UM\boldsymbol{\Lambda}\UM^\top\UM\boldsymbol{\Lambda}\UM^\top = \UM\boldsymbol{\Lambda}^2\UM^\top\)
for common neighbors, which accords with the function $f(\cdot)$ in Table~\ref{tab:linkpred}.
As for the $k$-hop RW, we have
$\hat{\AM}^k = \UM\boldsymbol{\Lambda}\UM^\top\cdots \UM\boldsymbol{\Lambda}\UM^\top = \UM\boldsymbol{\Lambda}^k\UM^\top$.
Based thereon, it can be shown
\begin{small}
\begin{align*}
\textstyle \sum_{k=0}^{\infty}{(1-\alpha)\alpha^k \hat{\AM}^k} & \textstyle= \sum_{k=0}^{\infty}{(1-\alpha)\alpha^k \UM\boldsymbol{\Lambda}^k\UM^\top}\\
& \textstyle = \UM \cdot \sum_{k=0}^{\infty}{(1-\alpha)\alpha^k \boldsymbol{\Lambda}^k}\cdot \UM^\top = \UM \frac{(1-\alpha)}{1-\alpha\boldsymbol{\Lambda}} \UM^\top.
\end{align*}
\end{small}
and
\begin{small}
\begin{align*}
\textstyle \sum_{k=0}^{\infty}{\frac{e^{-\alpha}\alpha^k}{k!} \hat{\AM}^k} & \textstyle = \sum_{k=0}^{\infty}{\frac{e^{-\alpha}\alpha^k}{k!} \UM\boldsymbol{\Lambda}^k\UM^\top} = \UM \sum_{k=0}^{\infty}{\frac{e^{-\alpha}\alpha^k}{k!} \boldsymbol{\Lambda}^k} \UM^\top = \UM \frac{e^{\alpha \boldsymbol{\Lambda}}}{e^\alpha} \UM^\top,
\end{align*}
\end{small}
which completes the proof.
\end{proof}

\begin{proof}[\bf Proof of Lemma~\ref{lem:curve-fitting}]
The root mean squared
error in the optimization objective $\min_{f^{(++)}\in \mathcal{F}}{\|f^{(++)}(\hat{\AM}^{+})-\YM^{+}\|_F}$ in Eq.~\eqref{eq:pos-pos-neg-pos} can be rewritten as:
\begin{small}
\begin{align*}
\|f^{(++)}(\hat{\AM}^{+})-\YM^{+}\|_F &= \|f^{(++)}({\UM}^{+}{\boldsymbol{\Lambda}}^{+}{{\UM}^{+}}^{\top})-\YM^{+}\|_F \\
& = \|{\UM}^{+}f^{(++)}({\boldsymbol{\Lambda}}^{+}){{\UM}^{+}}^{\top}-\YM^{+}\|_F\\
& = \|f^{(++)}({\boldsymbol{\Lambda}}^{+})-{{\UM}^{+}}^{\top}\YM^{+}{\UM}^{+}\|_F,
\end{align*}
\end{small}
which can be decomposed into the sum of squares of its diagonal entries of $f^{(++)}({\boldsymbol{\Lambda}}^{+})-{{\UM}^{+}}^{\top}\YM^{+}{{\UM}^{+}}$, and into the sum of squares of its off-diagonal entries that are independent of $f^{(++)}$~\cite{kunegis2009learning}.
The minimization of ${\|f^{(++)}(\hat{\AM}^{+})-\YM^{+}\|_F}$ is equivalent to optimizing the following least-squares curve fitting problem:
\begin{small}
\begin{equation*}
\textstyle \min_{f^{(++)}\in \mathcal{F}}{\sum_{i=1}^{|\U|+|\V|}{(f^{(++)}({\boldsymbol{\Lambda}_{i,i}}^{+})-{{\UM}_{\cdot,i}^{+}}^{\top}\YM^{+}{\UM}_{\cdot,i}^{+})^2}},
\end{equation*}
\end{small}
which finishes the proof.
\end{proof}

\begin{proof}[\bf Proof of Theorem~\ref{thm:gegenbauer}]
As stated by Theorem 3.21 in~\cite{shen2011orthogonal}, using the Rodrigues’ formula and integration by parts, when $K\in \mathbb{N}_0$ and $\alpha^\prime, \beta^\prime, a^\prime, b^\prime > -1$, the Jacobi polynomial basis~\cite{suetin2001ultraspherical} $P^{\alpha^\prime,\beta^\prime}_K(\lambda)$ can be expressed by \(P^{\alpha^\prime,\beta^\prime}_K(\lambda) = \sum_{k=0}^K{\hat{c}_k^K P^{a^\prime,b^\prime}_{k}(\lambda)}\),
where $\hat{c}_k^K$ is defined as
\begin{small}
\begin{align*}
\hat{c}_k^K = & \textstyle \frac{\Gamma(K+\alpha^\prime+1)}{\Gamma(K+\alpha^\prime+\beta^\prime+1)}\frac{(2k+a^\prime+b^\prime+1)\Gamma(k+a^\prime+b^\prime+1)}{\Gamma(k+a^\prime+1)} \\
& \textstyle \times \sum_{i=0}^{K-k}{\frac{(-1)^i\Gamma(K+k+i+\alpha^\prime+\beta^\prime+1)\Gamma(i+k+a^\prime+1)}{i!(K-k-i)!\Gamma(k+i+\alpha^\prime+1)\Gamma(i+2k+a^\prime+b^\prime+2)}}
\end{align*}
\end{small}
and $\Gamma(\cdot)$ stands for the Gamma function.

Since the Gegenbauer polynomial is a special case of the Jacobi polynomial , i.e., $J^\alpha_K(\lambda)=\frac{(2\alpha)_K}{(\alpha+\frac{1}{2})_K}\cdot P^{\alpha-\frac{1}{2},\alpha-\frac{1}{2}}_K(\lambda)$, we can derive that \(J^{\alpha}_K(\boldsymbol{\Lambda})=\sum_{k=0}^K{c_k^K\cdot J^{\alpha^\prime}_k(\boldsymbol{\Lambda})}\). Let \( (x)_K:=\prod_{j=0}^{K-1}{(x+j)} \) denote the Pochhammer symbol for $x\in \mathbb{C}$. Accordingly,
\( c_k^K = \frac{(2\alpha)_K}{(\alpha+\frac{1}{2})_K}\cdot \hat{c}_k^K\)
and $\alpha^\prime=\beta^\prime=\alpha-\frac{1}{2}$ and $a^\prime=b^\prime=a-\frac{1}{2}$. The theorem is proved.
\end{proof}

\begin{proof}[\bf Proof of Lemma~\ref{lem:svd-eig}]
Let $\boldsymbol{\alpha}^{(l)}\boldsymbol{\Sigma}{\boldsymbol{\alpha}^{(r)}}^\top$ be the SVD of $\BM$ and $\hat{\PsiM}\boldsymbol{\Lambda}\hat{\PsiM}^\top$ be the eigendecomposition of $\BM\BM^{\top}$. According to the relation between SVD and eigendecomposition~\cite{saad2011numerical}, i.e.,
\begin{equation*}
\BM\BM^{\top} = \boldsymbol{\alpha}^{(l)}\boldsymbol{\Sigma}{\boldsymbol{\alpha}^{(r)}}^\top\boldsymbol{\alpha}^{(l)}\boldsymbol{\Sigma}{\boldsymbol{\alpha}^{(r)}}^\top = \boldsymbol{\alpha}^{(l)}\boldsymbol{\Sigma}^2{\boldsymbol{\alpha}^{(r)}}^\top = \hat{\PsiM}\boldsymbol{\Lambda}\hat{\PsiM}^\top,
\end{equation*}
the columns in ${\alpha}^{(l)}$ are the eigenvectors of $\BM\BM^{\top}$ and the diagonal entries in $\boldsymbol{\Sigma}^2$ are the eigenvalues of $\BM\BM^{\top}$. 
Recall that the singular values are non-negative and sorted, i.e., $\boldsymbol{\Sigma}_{1,1} \ge \boldsymbol{\Sigma}_{2,2}\ge \ldots \ge \boldsymbol{\Sigma}_{|\U|+|\V|,|\U|+|\V|}$.
Hence, the diagonal entries in $\boldsymbol{\Sigma}^2$ are also non-negative and sorted, i.e., ${\boldsymbol{\Sigma}_{1,1}}^2 \ge {\boldsymbol{\Sigma}_{2,2}}^2\ge \ldots \ge {\boldsymbol{\Sigma}_{|\U|+|\V|,|\U|+|\V|}}^2$. In turn, the top-$k$ eigenvalues of $\BM\BM^{\top}$ are thus ${\boldsymbol{\Sigma}_{1,1}}^2,{\boldsymbol{\Sigma}_{2,2}}^2,\ldots,{\boldsymbol{\Sigma}_{k,k}}^2$, whose corresponding eigenvectors are $\boldsymbol{\alpha}^{(l)}_{\cdot,1},\boldsymbol{\alpha}^{(l)}_{\cdot,2},\ldots,\boldsymbol{\alpha}^{(l)}_{\cdot,k}$, which are exactly the top-$k$ left singular vectors $\boldsymbol{\alpha}$ of $\BM$. The lemma is hence proved.
\end{proof}


\section{Closed‐Form Expressions of $J^{\alpha}_k(\hat{\AM}^)$
}\label{eq:closed-form}


We define the Pochhammer symbol as \( (x)_k:=\prod_{j=0}^{k-1}{(x+j)} \) for $x\in \mathbb{C}$ and $k\in \mathbb{N}_0$. According to Chapter 2.1 in \cite{Reimer2003}, if $\alpha\neq 0$ and $k\in \mathbb{N}_0$, the Gegenbauer polynomial basis can be expressed by
\begin{small}
\begin{equation}
\textstyle J^{\alpha}_k(\lambda) = \sum_{i=0}^{\lfloor \frac{k}{2}\rfloor}{(-1)^i\frac{(\alpha)_{k-i}}{(1)_i(1)_{k-2i}}\cdot (2\lambda)^{k-2i}}.
\end{equation}
\end{small}
Therefore,
\begin{small}
\begin{align*}
\textstyle J^{\alpha}_k(\hat{\AM}^+) & \textstyle = \UM^+\cdot J^{\alpha}_k(\boldsymbol{\Lambda}^+)\cdot {\UM^+}^\top = \UM^+ \sum_{i=0}^{\lfloor \frac{k}{2}\rfloor}{(-1)^i\frac{(\alpha)_{k-i}}{(1)_i(1)_{k-2i}} (2\boldsymbol{\Lambda}^+)^{k-2i}} {\UM^+}^\top \\
&\textstyle =  \sum_{i=0}^{\lfloor \frac{k}{2}\rfloor}(-1)^i\frac{(\alpha)_{k-i}}{(1)_i(1)_{k-2i}}2^{k-2i}\cdot {\hat{\AM}^{+^{k-2i}}}.
\end{align*}
\end{small}
In the same vein,
\( J^{\alpha}_k(\hat{\AM}^-) 
 =  \sum_{i=0}^{\lfloor \frac{k}{2}\rfloor}(-1)^i\frac{(\alpha)_{k-i}}{(1)_i(1)_{k-2i}}2^{k-2i}\cdot {\hat{\AM}^{-^{k-2i}}}\).

\balance

\section{GenAI Usage Disclosure}
The authors confirm that no Generative AI tools were used in any part of this research, including data collection, code development, analysis, or manuscript writing.


\bibliographystyle{ACM-Reference-Format}
\bibliography{main}

\end{document}